\documentclass[acmtog,prologue,table]{acmart}

\usepackage{multirow}
\usepackage{subcaption}
\usepackage{dblfloatfix} 
\usepackage{nicematrix}
\usepackage{xcolor}
\usepackage{tcolorbox}

\begin{document}

\title{SonicDiffusion: Audio-Driven Image Generation and Editing with Pretrained Diffusion Models}

\author{Burak Can Biner}
\email{bbiner21@ku.edu.tr}
\orcid{}
\affiliation{%
  \institution{Koç University}
  \country{Turkey}
}

\author{Farrin Marouf Sofian}
\email{}
\orcid{}
\affiliation{%
  \institution{Koç University}
  \country{Turkey}
}

\author{Umur Berkay Karakaş}
\email{}
\orcid{}
\affiliation{%
  \institution{Koç University}
  \country{Turkey}
}

\author{Duygu Ceylan}
\email{duygu.ceylan@gmail.com}
\affiliation{%
  \institution{Adobe Research}
  \country{United Kingdom}
}

\author{Erkut Erdem}
\email{erkut@cs.hacettepe.edu.tr}
\orcid{0000-0002-6744-8614}
\affiliation{%
  \institution{Hacettepe University}
  \country{Turkey}
}

\author{Aykut Erdem}
\email{aerdem@ku.edu.tr}
\orcid{0000-0002-6280-8422}
\affiliation{%
  \institution{Koç University}
  \country{Turkey}
}

\renewcommand{\shortauthors}{Biner, et al.}

\begin{teaserfigure}
  \includegraphics[width=\textwidth]{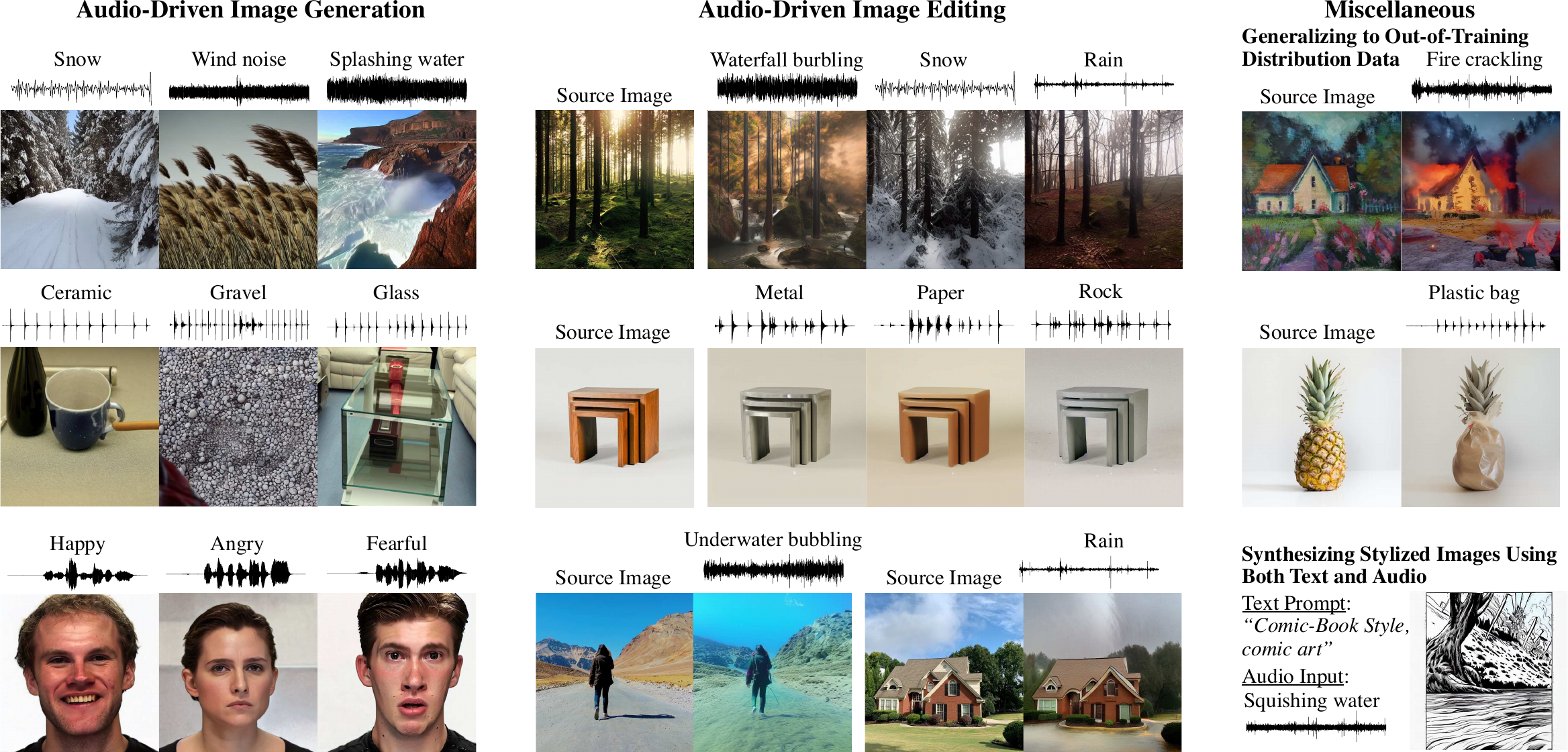}
  \caption{\textbf{SonicDiffusion}. Our framework introduces a novel approach where audio inputs guide image generation and editing. Leveraging paired audio-visual data, SonicDiffusion effectively learns to translate sounds into corresponding audio tokens and use them to guide the diffusion process. Beyond this, our method can combine audio with textual information for richer image synthesis, and is capable of artistically altering paintings, demonstrating its versatility and creative potential.}
  \Description{Description here.}
  \label{fig:teaser}
\end{teaserfigure}

\begin{abstract}
       We are witnessing a revolution in conditional image synthesis with the recent success of large scale text-to-image generation methods. This success also opens up new opportunities in controlling the generation and editing process using multi-modal input. While spatial control using cues such as depth, sketch, and other images has attracted a lot of research, we argue that another equally effective modality is audio since sound and sight are two main components of human perception. Hence, we propose a method to enable audio-conditioning in large scale image diffusion models. Our method first maps features obtained from audio clips to tokens that can be injected into the diffusion model in a fashion similar to text tokens. We introduce additional audio-image cross attention layers which we finetune while freezing the weights of the original layers of the diffusion model. In addition to audio conditioned image generation, our method can also be utilized in conjuction with diffusion based editing methods to enable audio conditioned image editing. We demonstrate our method on a wide range of audio and image datasets. We perform extensive comparisons with recent methods and show favorable performance.
\end{abstract}



\maketitle

\vspace{-0.25cm}
\section{Introduction}
\label{sec:intro}

As visual content synthesis evolves, the quest for more immersive and authentic generation techniques becomes paramount. Text-driven approaches like DALL·E 2 \cite{dalle2}, Latent Diffusion Models (LDM) \cite{ldm}, and Imagen \cite{imagen}, have shown remarkable results in terms of image realism and creativity. However, the reliance on text for image conditioning, while effective, introduces limitations due to the inherently manual and sometimes incongruent nature of textual descriptions with their visual counterparts. This observation underscores the potential of alternative modalities that offer a more natural and cohesive integration with visual content.

The integration of audio cues into the image generation process represents an intriguing yet underexplored frontier. Audio, unlike text, shares a direct and natural correlation with visual scenes, providing a rich tapestry of information that text might overlook or find cumbersome to articulate. Consider the complex soundscape of a bustling city street: audio not only captures the cacophony of voices, vehicle noises, and the distant hum of urban life but also conveys the atmosphere of the scene, from the brisk energy of a morning commute to the subdued tones of an evening stroll. Such auditory cues offer a depth of context and ambiance that text descriptions may struggle to encapsulate fully. By leveraging these rich audio signals, our approach aims to generate images that reflect the vibrancy and dynamism of visual content, showing the potential of audio to provide a more nuanced and comprehensive understanding of the scenes. To address these challenges, we propose SonicDiffusion, a new diffusion model that can generate images semantically aligned with accompanying audio, as illustrated in Fig.~\ref{fig:teaser}.

Prior efforts in audio-driven image synthesis, such as those based on Generative Adversarial Networks (GANs)~\cite{sgsim,robous_sgsim,into-the-wild} and pre-trained text-to-image models~\cite{gluegen,audiotoken} have laid the groundwork for our research. These studies have demonstrated the potential of audio to guide image generation. However, they often fall short in sample quality or struggle with limitations in the depth of audio's semantic capture. SonicDiffusion advances beyond these initial efforts by integrating audio cues within a diffusion model framework, specifically adapting and extending the capabilities of the Stable Diffusion model \cite{ldm} to embrace audio information.

Our approach is distinguished by its ability to process and translate audio signals into visual representations, employing audio-image cross-attention layers that serve as a conduit for this modality translation. This method not only maintains the high-quality image generation synonymous with diffusion models but also introduces a novel dimension of audio-driven creativity and contextual richness with minimal additional training requirements. Furthermore, our method can be easily extended to edit real images, performing modifications that reflect the characteristics of the audio input through feature injection.

We validate the efficacy of our approach through rigorous testing on three diverse datasets. Our results show that our model surpasses current existing methods for audio-to-image generation and audio-driven image editing, both in qualitative and quantitative terms. Notably, our model exhibits exceptional ability in capturing intricate high-frequency details in landscape generation, accurately rendering human facial features influenced by sound, and precisely mapping specific sounds to corresponding visual elements that reflect the material properties of depicted objects.

To sum up, our contributions are threefold: (i) We introduce a new diffusion model that extends the capabilities of the pre-trained Stable Diffusion model, enabling sound-guided image generation. (ii) Our approach presents audio-image cross-attention layers, strategically designed to require a minimal set of trainable parameters. (iii) We demonstrate the versatility of our model in not just generating, but also editing images to match auditory inputs. For our code and models, please see \url{https://cyberiada.github.io/SonicDiffusion/}.
\section{Related Work}
\label{sec:related_work}

\textbf{Diffusion-based Image Generation and Editing.}
Image generation and editing have evolved significantly with diffusion models like Imagen \cite{imagen}, DALL$\cdot$E 2 \cite{dalle2}, and Latent Diffusion \cite{ldm}, which generate realistic images from text prompts. These models are also central in text-based image editing~\cite{sdedit, GLIDE}. Various approaches have been developed since then. PnP \cite{plugandplay} injects image features and manipulates self-attention weights during text guided editing. Imagic~\cite{imagic} optimizes text embeddings and fine-tunes the model for targeted edits. Prompt-to-Prompt \cite{prompt2prompt} adjusts cross-attention maps to ensure spatial consistency during editing. InstructPix2Pix \cite{instructpix2pix} employs GPT-3 \cite{gpt-3} to generate editing instructions which are then used to train a diffusion model to enable instruction based editing. Pix2pix-zero \cite{pix2pix_zero} calculates edit directions from the text embeddings of original and edited image pairs. Lastly, \citet{null_text_inversion} focus on optimizing null-text embeddings to facilitate better inversion and editing performance. All these methods, however, focus on text guided generation and editing while our goal is to unlock audio guidance.

\noindent\textbf{Adapters.}
Adapters have proven to be highly effective for transfer learning within large pre-trained NLP models, achieving results that closely approach the state-of-the-art~\cite{peft}. For instance, adapters have recently been utilized in~\cite{llamaadapter} to tune LLaMa \cite{llamaadapter} into an instruction model. In the text-to-image diffusion models, adapters have also been instrumental. T2I-Adapter \cite{t2iadapter} illustrates that adapters can offer a straightforward, cost-effective, and flexible means of guiding pre-trained Stable Diffusion models while preserving their core structure.  IP-Adapter \cite{ye2023ipadapter} introduces an adapter architecture with a decoupled cross-attention mechanism, facilitating multimodal image generation.

\begin{figure*}[!t]
  \includegraphics[width=0.99\textwidth]{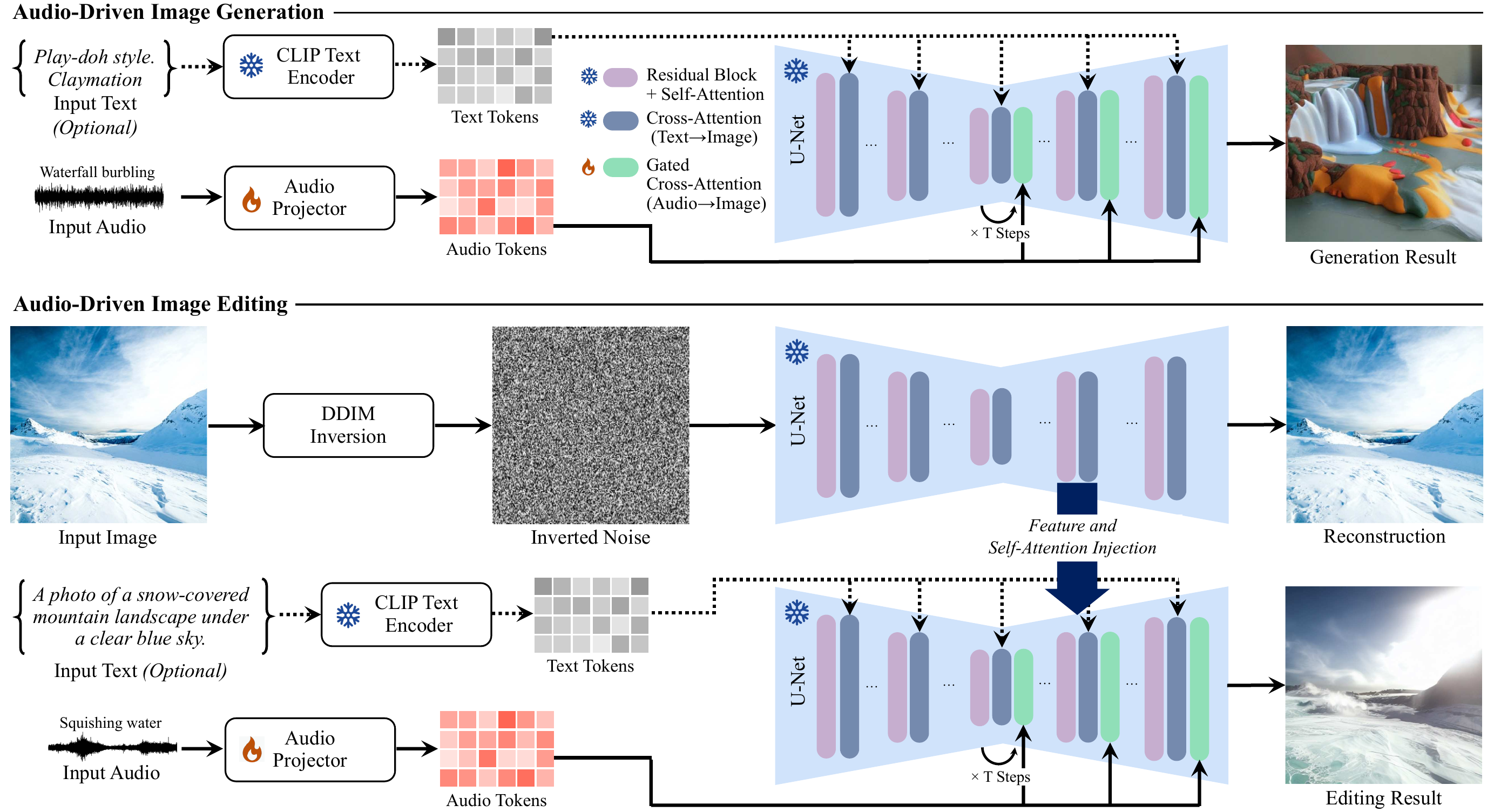}\vspace{-0.05cm}
  \caption{\textbf{An overview of our proposed SonicDiffusion model}. Our framework allows for two core functionalities: (1) audio-driven image generation, and (2) audio-guided image editing. In (1), both sound inputs and optional text prompts are tokenized, guiding the denoising process via text-to-image and audio-to-image cross-attention layers. For (2), the process begins with the inversion of the input image using DDIM. Subsequently, extracted spatial features and self-attention maps are integrated into the generation process, complemented by audio-conditioned cross-attention maps to obtain the desired changes. \vspace{-0.05cm}
  }
  \label{fig:inference_overview}
\end{figure*}

\noindent\textbf{Audio-Driven Image Generation and Editing.}
Incorporating audio into image generation and editing has recently gained traction. Early studies predominantly utilized GANs~\cite{gan}. SGSIM~\cite{sgsim} uses a two-step approach that augments the CLIP~\cite{clip} embedding space with audio embeddings via InfoNCE loss~\cite{infonce,representation}, combining audio, image, and text. It then manipulates StyleGAN's~\cite{stylegan} latent codes to produce images compatible with the audio inputs. Robust-SGSIM enhances this method by adding a KL divergence term to its loss function to better preserve image structure and reduce bias. Sound2Scene~\cite{sound2visual} also utilizes InfoNCE loss but combines it with BigGAN~\cite{brock2019large} for synthesis. AVStyle~\cite{into-the-wild} introduces an audio-visual adversarial loss, focusing on texture stylization and structural preservation using patch-wise contrastive learning.

Diffusion-based methods have gained popularity recently. GlueGen \cite{gluegen} is a flexible model handling multimodal inputs, merging linguistic contexts via XLM-R \cite{xlm-roberta} and auditory elements through AudioCLIP \cite{audioclip}, aligning representations within the CLIP framework to enable conditional image generation via \cite{ldm}. AudioToken \cite{audiotoken} presents an audio embedder that replaces a specific token's CLIP embedding, enabling audio-guided image generation. ImageBind \cite{imagebind} combines audio, image, and text modalities in one embedding space using InfoNCE, also enabling audio-driven image synthesis. \citet{lee2023generating} introduce a two-stage method, initially using an audio captioning transformer for generating attention map, followed by direct sound optimization to generate new images from the initial latent embeddings. Our model distinguishes itself from these methods by adaptively tuning the denoising process of the Stable Diffusion. In particular, we incorporate audio-image cross-attention layers to enhance the semantic and visual alignment between the synthesized images and the input audio.

\begin{figure*}[!h]
  \includegraphics[width=0.99\textwidth]{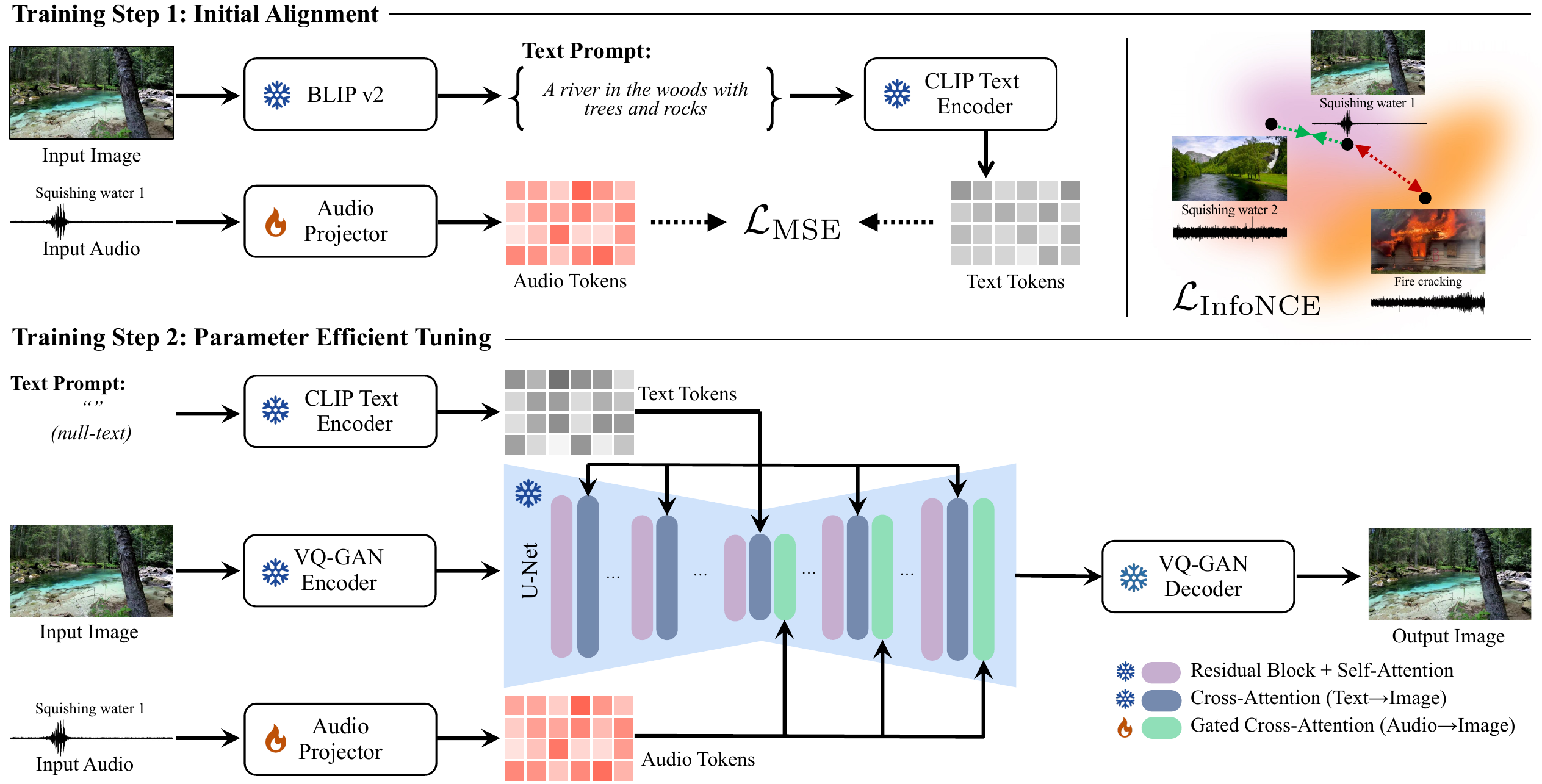}
  \centering\vspace{-0.05cm}
  \caption{\textbf{Training pipeline of SonicDiffusion} comprises two distinct stages: (1) aligning audio features with CLIP's semantic space, and (2) sound-driven parameter-efficient tuning of the Stable Diffusion (SD) model. In stage (1), the audio projector module is trained to transform audio clips into semantically rich tokens, employing MSE and contrastive loss functions. Stage (2) involves integrating gated cross-attention layers, which facilitate interaction between image and audio modalities, into the existing SD framework. Only these newly added layers, alongside the audio projector module, are adjusted to enable audio-conditioned image synthesis. 
  }
  \label{fig:train_overview}
\end{figure*}

\noindent\textbf{Audio-Driven Video Generation.}
 Recent years have seen growing interest in generating videos from audio. Building on their earlier work \cite{sgsim}, \citet{landscape} developed a method combining a sound inversion encoder with a StyleGAN generator, using recurrent blocks for fine texture variations in video frames. TPoS \cite{jeong2023power} similarly employs an audio encoder with recurrent blocks, adding a temporal attention module for better audio temporal understanding. This model uses the multimodal CLIP space from \cite{sgsim} for image synthesis. Another recent approach is TempoTokens \cite{yariv2023diverse}, which proposes to use specific tokens generated from audio using an audio mapper. The model leverages an audio-conditioned temporal attention mechanism and facilitates the generation of videos by integrating these tokens into a pre-trained text-to-video diffusion model.

\section{Method}
In this work, we introduce SonicDiffusion, an approach that steers the process of image generation and editing using auditory inputs. As depicted in Fig.~\ref{fig:inference_overview}, our proposed approach has two principal components. The first module, termed the {Audio Projector}, is designed to transform features extracted from an audio clip into a series of inner space tokens. These tokens are subsequently integrated into the image generation model through newly incorporated audio-image cross-attention layers. Crucially, we maintain the original configuration of the image generation model by freezing its existing layer weights. This positions the added cross-attention layers as adapters, serving as a parameter-efficient way to fuse the audio and visual modalities. Below, we first present preliminaries about the Stable Diffusion (SD) model, which serves as the basis for our implementation (Sec.~\ref{sec:pre}). We then detail the specifics  of the audio projector (Sec.~\ref{sec:proj}) and discuss the methodology employed for the integration of audio tokens into the SD framework (Sec.~\ref{sec:inj}). Furthermore, we describe how we can extend our approach to sound-guided image editing through feature injection (Sec.~\ref{sec:editing}).

\subsection{Preliminaries}
\label{sec:pre}
Stable Diffusion (SD) is a latent text-to-image diffusion model (LDM) \cite{ldm}, integrating a variational autoencoder to project an input image $\mathbf{x}$ into a reduced latent space $\mathbf{z} = E(\mathbf{x})$ via its encoder $E$. The corresponding decoder $D$ reconstructs an image $\hat{\mathbf{x}} = D(E(\mathbf{x}))$ from any given latent $\mathbf{z}$. Central to SD is its diffusion UNet \cite{unet}, operating in this latent space to progressively denoise input latents at each diffusion step. An unconditional UNet $\epsilon_{\theta}$ is trained with the established denoising loss introduced by Denoising Diffusion Probabilistic Models (DDPM) \cite{ddpm}:
\begin{equation}
    \mathcal{L}_{LDM} = \mathbb{E}_{\mathbf{z}\sim E(
\mathbf{x}), \epsilon \sim N(0, 1), t} \left[ ||\epsilon - \epsilon_{\theta} (\mathbf{z}_t, t) ||_{2}^{2} \right] 
\end{equation}
where $\mathbf{z}_t$ represents the noise corrupted latent at time step $t$, and $\mathbf{z}_0 = E(\mathbf{x})$.

The conditional variant of LDM incorporates a cross-attention mechanism, enabling it to be conditioned with different modalities. Specifically, SD leverages text conditioning through embeddings generated by the CLIP text encoder. Consequently, the denoising loss is adapted to accommodate this conditioning, as follows:
\begin{equation}
    \mathcal{L}_{LDM} = \mathbb{E}_{\mathbf{z}\sim E(\mathbf{x}), \epsilon \sim N(0, 1), t, \mathbf{y}} \left[ ||\epsilon - \epsilon_{\theta} (\mathbf{x}, t, c_{\phi}((\mathbf{y})) ||_{2}^{2} \right]
    \label{ddpm_condition_loss}
\end{equation}
where $\mathbf{y}$ is the conditioning text prompt, and $c_{\phi}$ is the text encoder transforming $\mathbf{y}$ into an intermediate representation.

\subsection{Training Pipeline}
\label{app:training-pipeline}
The overall training pipeline is illustrated in Fig.~\ref{fig:train_overview}. SonicDiffusion employs a two-stage training process, each serving a specific purpose in achieving audio-driven image synthesis. The initial stage is dedicated to training the audio projector module. The second stage of training focuses on integrating gated cross-attention layers within the SD model. Below, we discuss the details of data processing as well as the two-stage training framework.

\subsection{Audio Projector}
\label{sec:proj}
\begin{figure}[!t]
  \includegraphics[width=\linewidth]{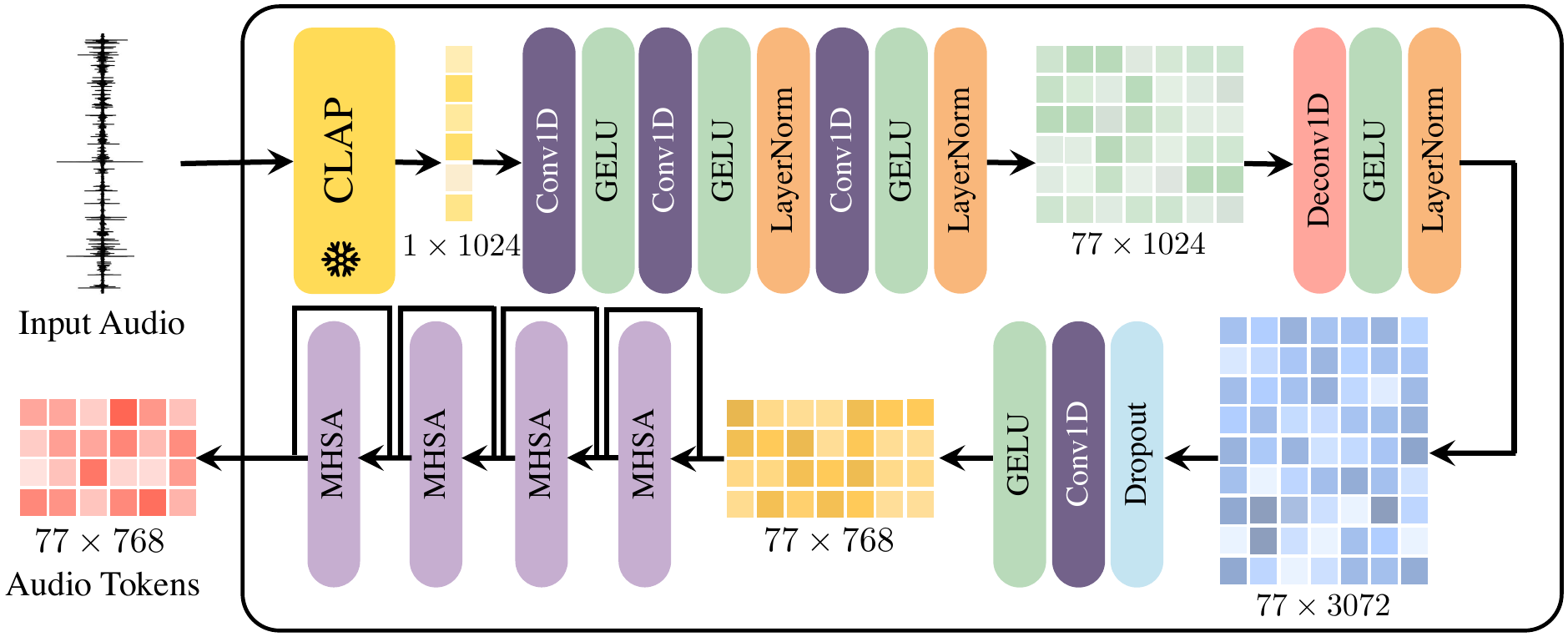}
  \caption{\textbf{Audio Projector} utilizes 
  CLAP audio encoder~\cite{clap} for initial feature extraction, followed by a mapper with 1D convolutions and deconvolutions. Four self-attention layers are then integrated to enhance learning effectiveness of the projector.}
 
  \label{audio_projector}
\end{figure}

In training the Stable Diffusion model, a large text-image pair dataset was used to learn a text-conditioning space semantically aligned with the image space. To achieve similar alignment between audio and image domains, we introduce the \emph{Audio Projector} module. This module, depicted in Fig.~\ref{audio_projector}, converts audio clips into tokens for additional conditioning input. It utilizes the CLAP model \cite{clap}, a state-of-the-art audio encoder, to generate one-dimensional embeddings from audio inputs. These embeddings are then transformed by an initial mapper with 1D convolutional and deconvolutional layers into $K$ tokens of $C$ channels. These tokens are further refined by four self-attention blocks to produce final representations, which are compatible with the cross-attention layers of the SD model, matching the dimensions of the text tokens.

The training of Audio Projector includes a joint strategy involving both Mean Squared Error (MSE) and contrastive loss functions.
Specifically, for the contrastive loss, we employ the InfoNCE \cite{infonce} loss, treating audio clips from the same class as positive pairs, and those from different classes as negative pairs. Within each training batch, two audio clips, $\mathbf{a}_0$ and $\mathbf{a}_1$, are randomly selected from the same class, alongside $N$ audio clips, denoted as $\left\{\textbf{a}_i\right\}_{i=2}^{N+1}$, from other classes. The contrastive loss objective is thus formulated as:
\begin{equation}
   \mathcal{L}_\text{InfoNCE} = -\log \frac{\exp(\langle \mathbf{a}_{0},\mathbf{a}_{1}\rangle)}{\sum_{i=1}^{N+1} \exp(\langle \mathbf{a}_{0},\mathbf{a}_{i}\rangle) }
\end{equation}
This contrastive loss pushes different classes away from each other, resulting in a well-defined semantic space. To enrich this space, we utilize audio-image pairs from training dataset. For each audio clip, captions are extracted from corresponding images using the BLIP v2 \cite{blipv2} and their text embeddings are computed using the text encoder of SD. An MSE loss is defined as the squared norm difference between the audio tokens $\mathbf{c}_\text{audio}$ and the text tokens $\mathbf{c}_\text{text}$:
\begin{equation}
\mathcal{L}_\text{MSE} =  \| \mathbf{c}_\text{audio} - \mathbf{c}_\text{text} \|_2^2,
\end{equation}
The training of the audio projector combines these two losses with weights $\alpha_1=1.0$ and $\alpha_2=0.25$, respectively:
\begin{equation}
    \mathcal{L}_{\text{AudioProjector}} = \alpha_1*\mathcal{L}_\text{InfoNCE} + \alpha_2*\mathcal{L}_\text{MSE} 
\end{equation}
\subsection{Gated Cross-Attention for Audio Conditioning}
\label{sec:inj}
\begin{figure}[!t]
  \includegraphics[width=\linewidth]{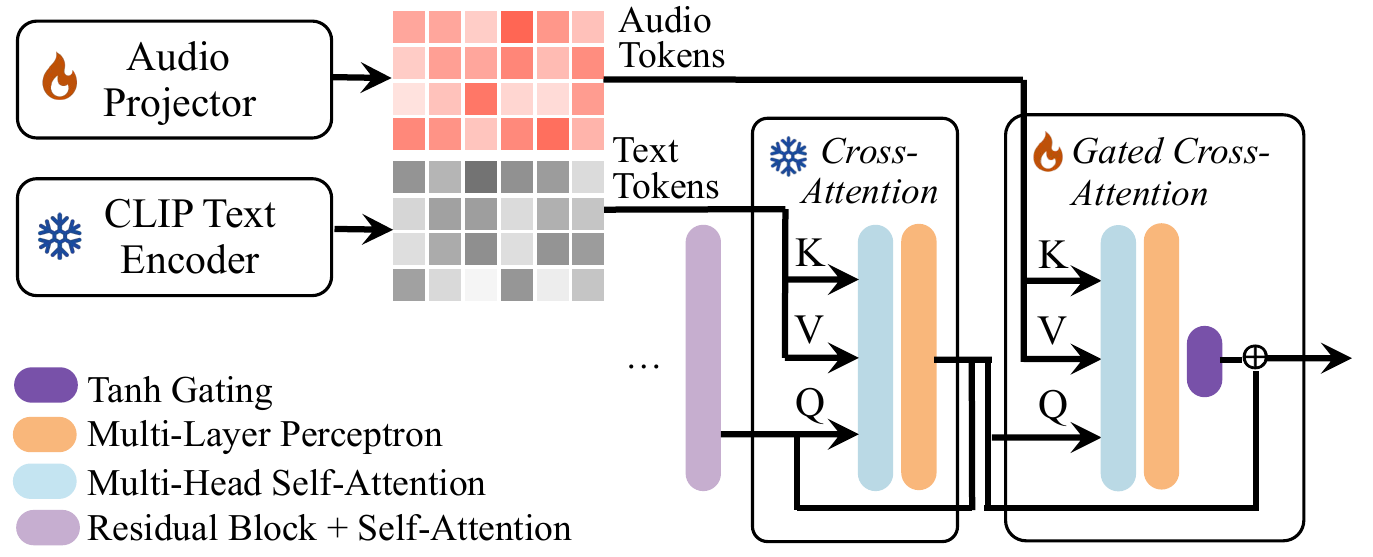}
  \centering
  \caption{\textbf{Gated cross-attention module.} Our method guides the diffusion process by gated cross-attention and dense feed-forward layers added after pre-trained text-conditioned layers, using audio tokens as keys/values and ensuring training stability and quality.
  }
  \label{gated_cross_attention}
\end{figure}

The diffusion UNet architecture of SD comprises blocks that include residual, self-attention, and cross-attention layers. The self-attention layers capture fine-grained spatial information, while the cross-attention layers establish semantic relationships between the conditioning text and the generated image. To effectively model the relationship between audio tokens and the image, we introduce new gated cross-attention layers into this architecture (Fig.~\ref{gated_cross_attention}). This modification aims to facilitate audio-conditioned image generation by enabling image features to interact dynamically with audio tokens.

We have retained the encoder component of the UNet, while augmenting the decoder with our newly introduced gated cross-attention layers. In this configuration, only the audio projector and these new layers are set as trainable, with the rest of the network remaining frozen. The pre-trained audio projector is further refined through DDPM loss optimization, enhancing the alignment between audio and image pairs. 
Within each decoder block of the UNet, we process visual tokens $
\mathbf{v}$, conditioning text tokens $\mathbf{c}_\text{text}$, and audio tokens $\mathbf{c}_\text{audio}$ as follows:
\begin{align}
\mathbf{v} = \mathbf{v} + \text{Residual\_Block}(\mathbf{v}),\\
\mathbf{v} = \mathbf{v} + \text{Self\_Attn}(\mathbf{v}),\\
\mathbf{v} = \mathbf{v} + \text{Cross\_Attn}(\mathbf{v},\mathbf{c}_\text{text}), \label{eq_cross_text}\\
\mathbf{v} = \mathbf{v} + \beta*\text{tanh}(\gamma)* \text{Cross\_Attn}(\mathbf{v},\mathbf{c}_\text{audio})\;.\label{eq_cross_audio}
\end{align}

A key novelty in our approach is the incorporation of a gating mechanism within these new layers. This mechanism is crucial for maintaining the original denoising efficiency of the model. We initialize the trainable parameter $\gamma=0$ and set $\beta=1$. During training, $\gamma$ is optimized to foster gradual learning. The conditioned DDPM loss objective is applied at this stage, as defined below:
\begin{equation}
    \mathcal{L}_{LDM} = \mathbb{E}_{\mathbf{z}
    \sim E(\mathbf{x}), \epsilon \sim N(0, 1), t, \mathbf{c}_\text{audio}} \left[ ||\epsilon - \epsilon_{\theta} (\mathbf{x}, t, \mathbf{c}_\text{audio}) ||_{2}^{2} \right]
    \label{ddpm_condition_loss}
\end{equation}
Note that while null-text is used during tuning, our approach can incorporate an optional text prompt at inference time to augment the generation process.

\subsection{Audio-Guided Image Editing}
\label{sec:editing}
By extending SD to perform audio conditioned image generation, our method can be combined with existing SD-based image editing methods to enable audio-guided image editing. We demonstrate this by using the Plug-and-Play features method \cite{plugandplay}. This approach involves injecting residual and self-attention features, derived from inverting a source image, into a subsequent image generation phase. This process ensures the preservation of the structure of the source image while facilitating text-conditioned edits. 

In line with this approach, we inject features from the residual layers $f_{t}^{4}$ and attention maps $A_{t}^{l}$, obtained during the inversion step $t$, into the corresponding editing phase as follows:
\begin{equation}
    \mathbf{z}_{t - 1} = \epsilon_{\theta} (\mathbf{x}_t, t, c_{\phi}(\mathbf{y}), {f_{t}^{4}}, {{A_{t}^{l}}})
\end{equation}

Through the injection of self-attention features, our model captures the relationship between spatial features, preserving critical layout and shape details. By incorporating audio tokens as conditioning signals, it effectively facilitates audio-guided image editing. 

\subsection{Implementation Details}
Our experiments were conducted on a single NVIDIA V100 GPU. In both stages, the AdamW optimizer with a weight decay of 1e-2 was utilized. During the first stage, the audio projector was trained using a learning rate of 1e-4. For the second stage, the cross-attention layers were trained at a learning rate of 1e-4, while the audio projector was adjusted to a lower rate of 1e-5. We used a batch size of 6. For our image generation backbone, we used Stable Diffusion version 1.4, the same backbone all our Stable-Diffusion based competitors employ, to ensure a fair comparison in our evaluations. We employed the DDIM sampler with 50 steps and set the classifier-free guidance scale as 8 for our evaluation setup.

\section{Experimental Setup}
\label{sec:experimental_setup}
\noindent\textbf{Datasets}. We carry out our experiments on the following datasets.

\noindent\textit{Landscape + Into the Wild.} For the landscape scenes, we create a dataset by combining videos from Into the Wild \cite{into-the-wild} and High Fidelity Audio-Video Landscape (Landscape) datasets~\cite{landscape}, both gathered from YouTube. The Into the Wild dataset contains 94 hiking videos with nature sounds, and the Landscape dataset is composed of 928 high-resolution videos of varying lengths. The collected videos are split into 10 seconds, and a frame is extracted from each interval along with its corresponding 10-second audio. The final dataset contains 22,000 image-audio pairs.

\noindent\textit{Greatest Hits~\cite{greatest_hits}} is a video dataset that contains distinctive sounds that come from various objects when a drumstick hits them. Every hit in these videos is annotated by which material they have been hit on. There are 17 material types, some of which are wood, metal, dirt, rock, leaf, plastic, cloth, and paper. In total, there are 977 videos with 46,577 hit actions. Similar to the Landscape dataset, we extract 10 seconds of audio clips and a frame from that interval to have matching audio and image pairs. However, during the recording of this dataset, some materials are being hit by turns in a random order within a couple of seconds. To have more consecutive audios, we filter actions that have one material consecutively being hit. The final dataset contains 3,385 image-audio pairs.

\noindent\textit{RAVDESS \cite{ravdess}} dataset consists of 7,356 song-video files of 24 male and female actors. The dataset includes speech samples representing emotions like calm, happy, sad, angry, fearful, surprise, and disgust. In our experiments, we took the song-video version of this dataset and extracted a single frame along with its corresponding song from each video since the videos were only 4-5 seconds long. The final dataset contains 1,008 frames with their paired audio.

\noindent\textbf{Evaluation Metrics}. In our study, we conduct a quantitative evaluation focusing on two critical aspects: (i) the semantic relevance of generated images to the input audio, and (ii) the photorealism of the samples.
To assess the semantic relevance, we employ three specific metrics: Audio-Image Similarity (AIS), Image-Image Similarity (IIS), and Audio-Image Content (AIC), as introduced in \cite{audiotoken}. For evaluating the sample quality, we utilize the FID metric~\cite{frechet_inception_distance}. 
The detailed definitions of these metrics are provided in the supplementary.

\definecolor{mycolor}{HTML}{DDF2FD} 
\definecolor{mycolor2}{HTML}{FDFD96}
\definecolor{pnpcolor}{HTML}{FEB5B1}

\begin{table}[!t]
\caption{\textbf{Quantitative comparison of our proposed approach with existing audio-conditioned image generation methods}, focusing on AIS, AIC, ISS for semantic consistency, and FID for image quality. Top scores are \textbf{bolded}, second best are \underline{underlined}. Pre-trained large-scale models and audio-to-video generation model are highlighted with $*$ and $\dag$ symbols, respectively.}
\resizebox{0.96\linewidth}{!}{
\begin{tabular}{llcccc} 
\toprule
& Model &AIS$\uparrow$&AIC$\uparrow$&IIS$\uparrow$&FID$\downarrow$\\
\midrule
\multirow{8}{*}{\rotatebox[origin=c]{90}{\small Landscape + Into the Wild}} 
& SGSIM~\cite{sgsim} & .7224 & .2166 & .6898 & 220.3 \\
& Sound2Scene~\cite{sound2visual} & \underline{.7466} & .3672 & \underline{.8894} & \underline{122.2}  \\
& GlueGen~\cite{gluegen} & .6632  & \underline{.4618} & .7357 & 133.0  \\
 & ImageBind~\cite{imagebind}$^*$ & .7209 & .4600 & .8044 & 159.1 \\
 & CoDi~\cite{codi}$^*$  & \textbf{.7578} &  .3618 &  .7749 &  134.6 \\
& AudioToken~\cite{audiotoken} & .6983  & .2851 & .8592 & 141.3  \\
 & TempoTokens~\cite{yariv2023diverse}$^\dag$ & .7446  & .2242 & .6215 & 258.0  \\
& SonicDiffusion (Ours) & .7390 & \textbf{.5436} & \textbf{.8898} & \textbf{118.6}  \\ 
\midrule
\multirow{6}{*}{\rotatebox[origin=c]{90}{\small Greatest Hits}}
& Sound2Scene~\cite{sound2visual} & \underline{.6536} & .3125 & .6693 & 143.1  \\ 
& SGSIM~\cite{sgsim} & .5065 & .2000 & .6612 & 239.2  \\
& GlueGen~\cite{gluegen} & .5047  & \underline{.5050} & .5976 & 208.9  \\
& ImageBind~\cite{imagebind}$^*$ &  .5736 &  .215 &  .5889 &  186.0  \\
& CoDi~\cite{codi}$^*$ &  .5694 &  .2325 &  .5721 &  218.0  \\
& AudioToken~\cite{audiotoken} & \textbf{.6552} & .2675 & \textbf{.7541} & \underline{123.6}  \\
& SonicDiffusion (Ours) & .6237 & \textbf{.6050} & \underline{.7411} & \textbf{123.5}  \\ 
\midrule
\multirow{6}{*}{\rotatebox[origin=c]{90}{\small RAVDESS}}
& Sound2scene~\cite{sound2visual} & 5053 & .1257 & 5301 & \underline{140.3} \\
& SGSIM~\cite{sgsim} & .5090 & .1108 & .5094 & 155.4  \\
& Gluegen~\cite{gluegen} & .5052  & \underline{.2252} & .5059 & 247.1  \\
& Imagebind$^*$ & \textbf{.5802} & \underline{.2131} & .6332 & 248.8  \\
& CoDi~\cite{codi}$^*$ & .5292 & .1396 &.5674 & 229.1  \\
& AudioToken~\cite{audiotoken} & .5009 & .1821 & \underline{.6409} & 279.4  \\
& SonicDiffusion (Ours) & \underline{.5309} & \textbf{.2316} & \textbf{.8736} & \textbf{89.6}  \\ 
\bottomrule
\end{tabular}
}
\label{table:generation_results}
\end{table}

\noindent\textbf{Competing Approaches}. 
Our SonicDiffusion model is compared against seven state-of-the-art methods: the GAN-based SGSIM~\cite{sgsim} and Sound2Scene~\cite{sound2visual}, Stable-Diffusion based GlueGen~\cite{gluegen}, CoDi~\cite{codi}, AudioToken~\cite{audiotoken}, and TempoTokens~\cite{yariv2023diverse}, and DALLE$\cdot$2-based ImageBind~\cite{imagebind}. Unlike other models, TempoTokens is designed for video generation from audio clips; thus, we use the center frames of its generated videos in our evaluation. SGSIM, GlueGen, AudioToken, and Sound2Scene are fine-tuned on our datasets, while pre-trained models were used for large CoDi and ImageBind models. No fine-tuning is needed for TempoTokens, as it is already trained on the Landscape dataset. Further model and fine-tuning details are in the supplementary material. We also compare our image editing results with the text-based PnP model~\cite{plugandplay}, using audio class labels as text prompts. This comparison shows the advantage of injecting audio-derived semantic information into image manipulation, as opposed to relying solely on text labels.

\section{Results}
\label{sec:results}

\begin{figure*}[!t]
  \includegraphics[width=0.99\textwidth]{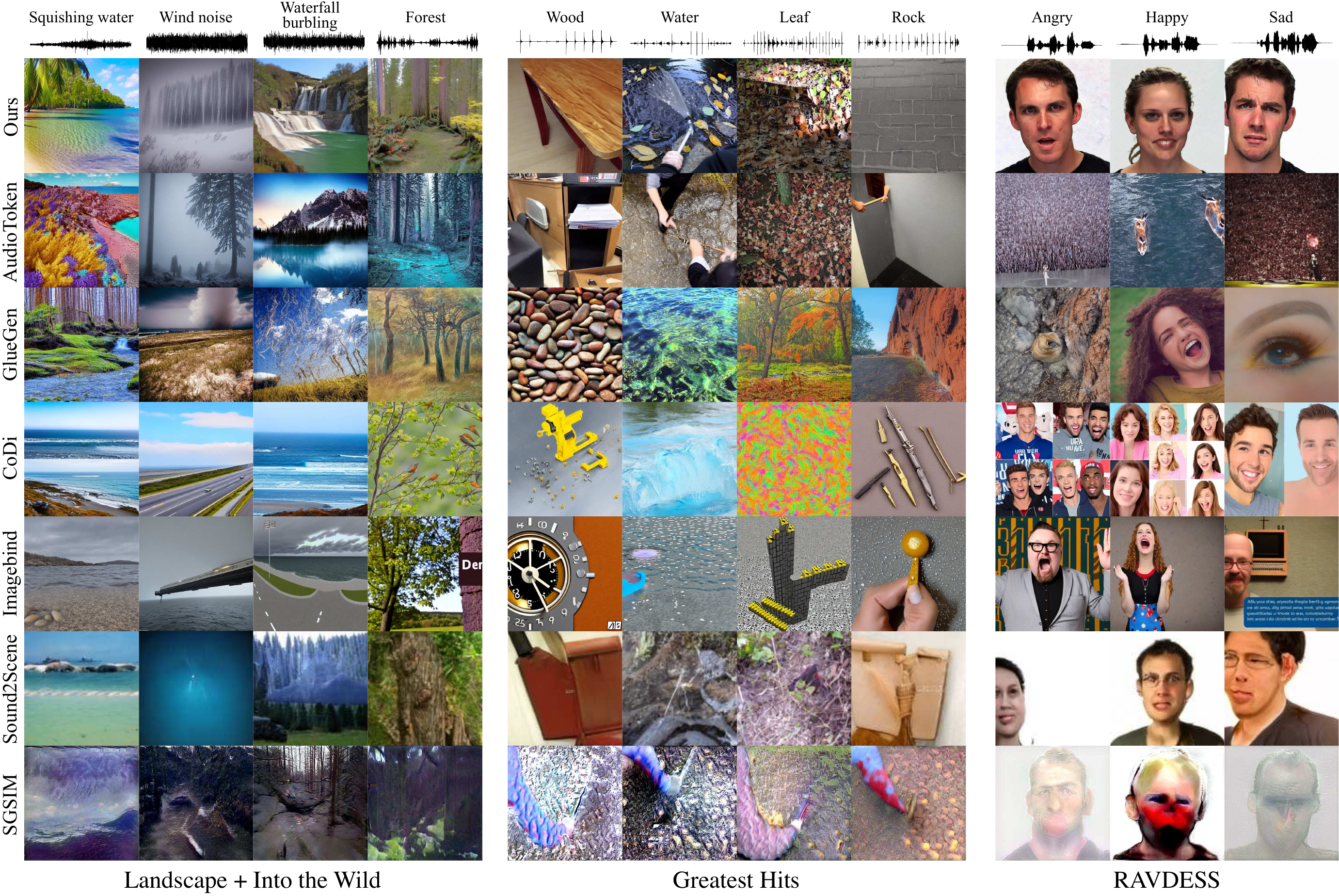}
  \caption{\textbf{Comparison against the state-of-the-art audio-driven image synthesis methods.} Our model generates images that closely align with the semantics of the input audio clips, surpassing all other existing methods in performance and fidelity.}
  \label{fig:generation_comparisons}
\end{figure*}

\noindent\textbf{Audio-Driven Image Generation.} Table~\ref{table:generation_results} shows quantitative evaluation results. Our model consistently outperforms competing approaches in photorealism, achieving the best FID scores across all datasets. Notably, our landscape images exhibit a good alignment with the audio semantics. In the Greatest Hits dataset, our model achieves the highest AIC score and the second-highest IIS score, further showing its effectiveness in audio-driven image generation.

In Fig.~\ref{fig:generation_comparisons}, we present a visual comparison between our model and the state-of-the-art methods. Our model demonstrates a superior capability in generating images that are not only visually coherent but also closely aligned with the input audio cues. For example, landscape images synthesized by our model capture the essence of the target scenes more accurately. When processing audio encoding material properties, our model synthesizes images featuring objects or elements that reflect the desired attributes. Additionally, the human face images generated from speech audio effectively mirror the corresponding vocal tonations, capturing the intended emotions.

\begin{figure}[!t]
    \centering
    \includegraphics[width=\linewidth]{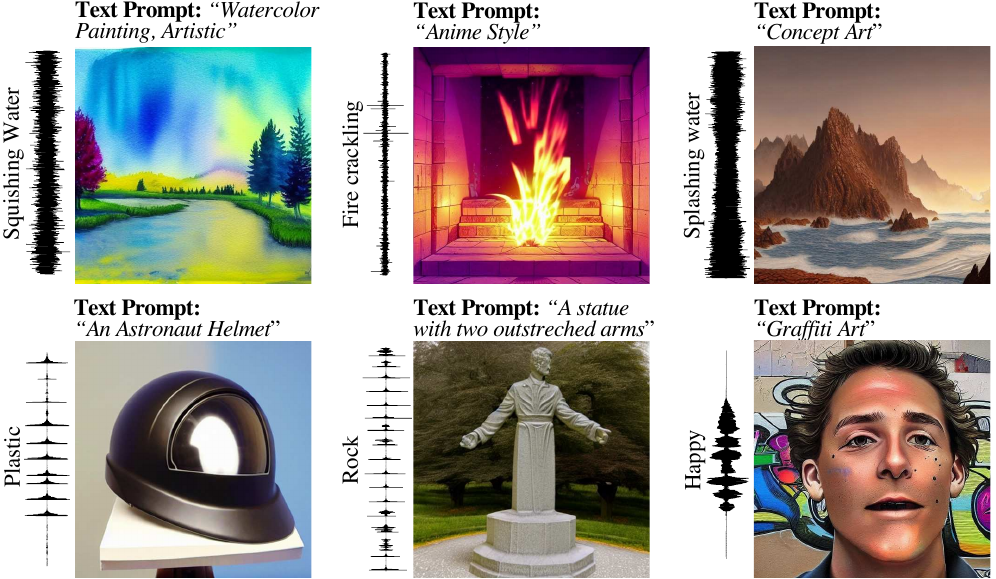}%
    \caption{\textbf{Image generation results using both text and audio.} Our model enables mixing text and audio modalities while synthesizing novel images.}
    \label{fig:mixed-generation}
\end{figure}

Moreover,  the versatility of our in handling mixed-modality inputs, blending both text and audio, is illustrated in Fig.~\ref{fig:mixed-generation}. These results demonstrate our method's proficiency in interpreting and fusing concepts from varied modalities. For example, it successfully merges a \textit{`watercolor painting'} text prompt with a \textit{`squishing water'} audio input, resulting in accurately composed and distinct image.

\setcounter{figure}{8}
\begin{figure*}[!b]
  \centering
  \includegraphics[width=0.975\textwidth]{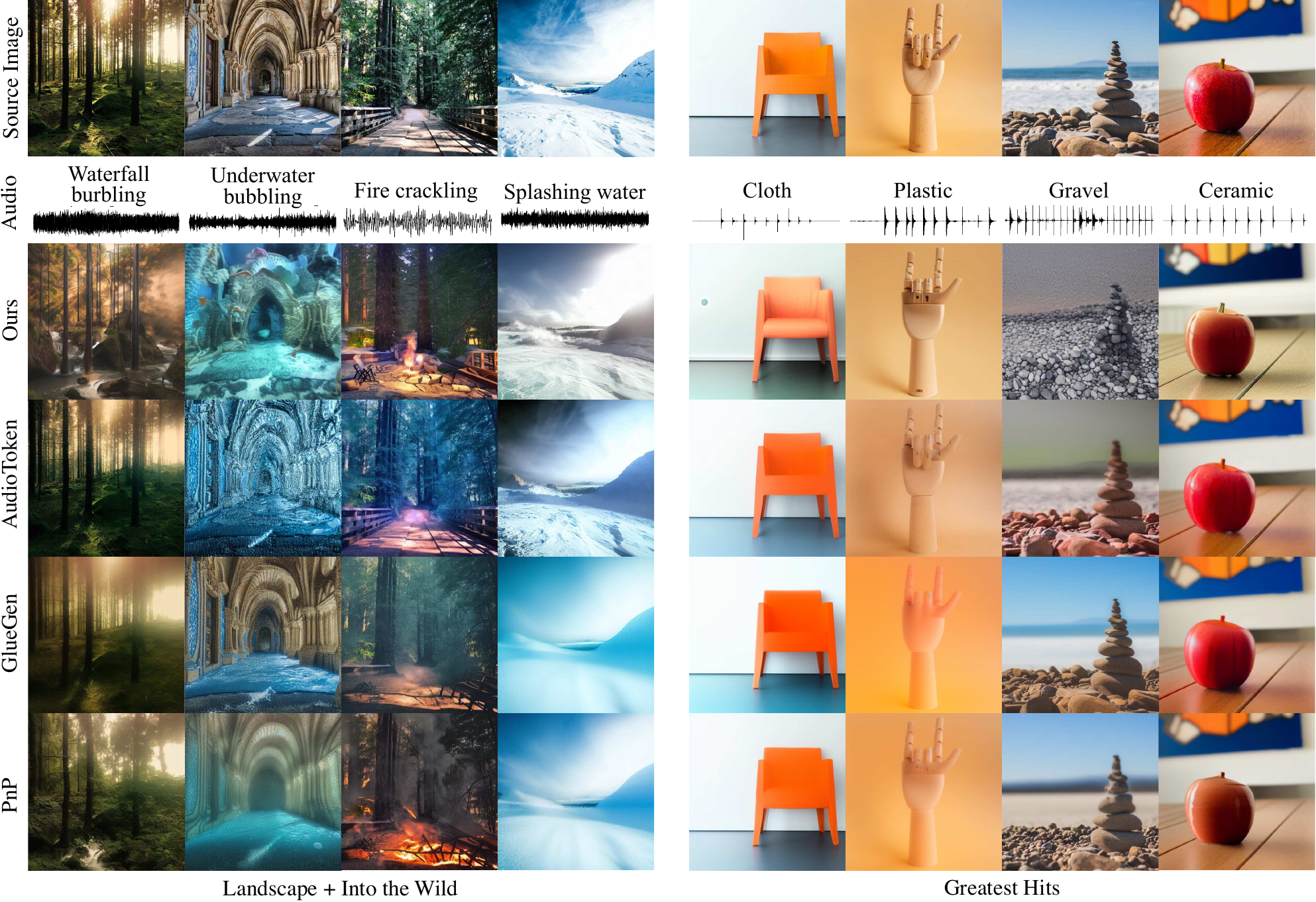}
  \caption{\textbf{Comparison against the state-of-the-art audio-driven image editing methods.} Our model generates images that closely align with the semantics of the input audio clips, surpassing all other existing methods in performance and fidelity.}
  \label{fig:editing_comparisons}
\end{figure*}

\setcounter{figure}{7}
\begin{figure}[!t]
    \centering
    \includegraphics[width=\linewidth]{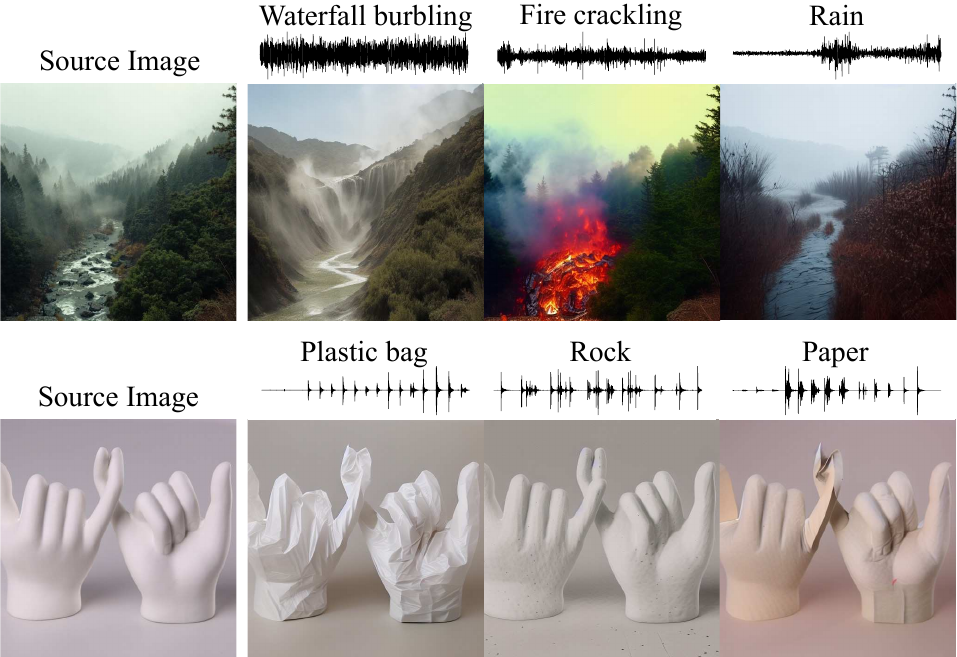}
    \caption{\textbf{Audio-guided image editing results.} Our model successfully performs manipulations that semantically reflect the audio content while maintaining the content of the source images.}
    \label{fig:one-source-many-results}
\end{figure}
\setcounter{figure}{9}



\noindent\textbf{Audio-Driven Image Editing.}
Our model also excels in editing real images based on audio clips. Fig.~\ref{fig:one-source-many-results} illustrates various examples of our method's image editing results. These show our model's ability in modifying image content to align with the semantics of the provided audio. For example, when editing a landscape image with the sounds of \textit{`waterfall burbling'}, \textit{`fire cracking'} and \textit{`rain'}, our method makes the necessary changes in the overall look of the input image. In another instance, as shown in the second row, the model successfully modifies the visual appearance of the objects based on the audio inputs reflecting material characteristics like \textit{`plastic bag'}, \textit{`rock'} and \textit{`paper'}. 

In Fig.~\ref{fig:editing_comparisons}, we show a comparative analysis of our model's editing capabilities against existing methods, including the text-based image editing PnP model, which serves as a foundational component of our framework. For this comparison, we employ class label information from audio inputs as text prompts in the PnP model. The results show that SonicDiffusion surpasses others in manipulation quality and style transfer accuracy. For instance, in editing the second image, methods like AudioToken and GlueGen not only fall short in achieving convincing manipulations but also introduce artifacts and color shifts. Similarly, the text-based PnP model fails to capture the full extent of the modifications implied by the audio inputs, unlike our  model.

\begin{figure*}[!b]
\centering
  \includegraphics[width=\linewidth]{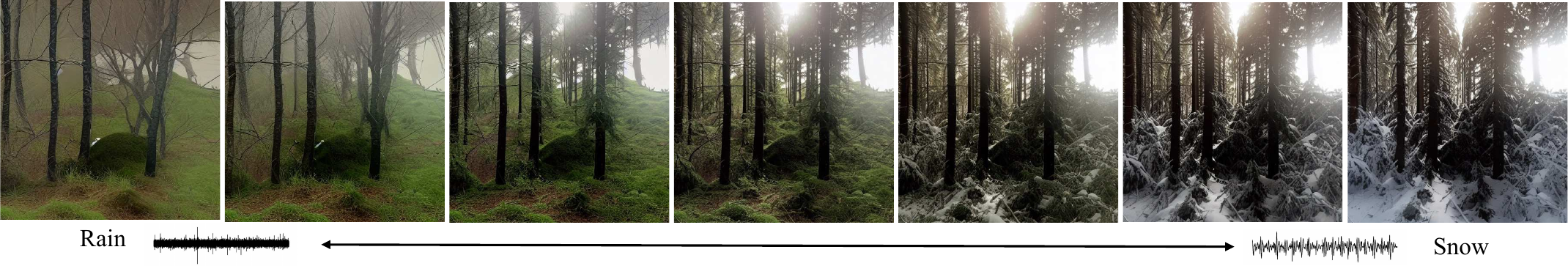} \vspace{-0.4cm}\\
  \small{(a) Sound mixing}\vspace{0.2cm}\\
  \includegraphics[width=\linewidth]{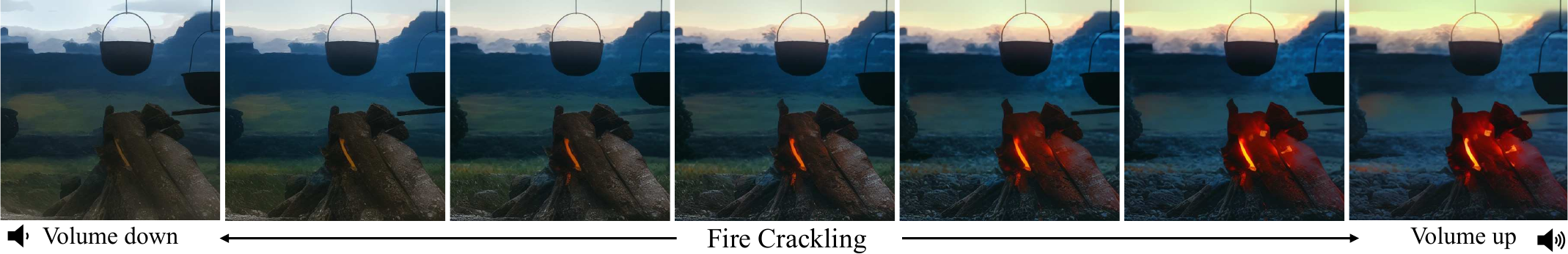} \vspace{-0.3cm}\\
 \small{(b) Volume changes}\vspace{-0.2cm}
 \caption{\textbf{Controlling Editing with Sound Manipulation:} Our model can (a) blend two sounds via linear interpolation of audio embeddings, and (b) adjust manipulation intensity by varying sound volume.}
 \label{fig:sound_manipulation}
\end{figure*}

Quantitative results in Table~\ref{tab:editing} further validate our approach. SonicDiffusion achieves the highest scores in AIS and ISS metrics, and the second-highest in IIS, confirming its ability to accurately manipulate images in response to audio cues. It also leads in FID scores, indicating better image quality compared to existing approaches. In a user study, our results are preferred by a majority of participants over competing approaches. Specifically, 56.05\% of 19 participants favored our method over GlueGen (17.37\%) and AudioToken (26.58\%) for the Greatest Hits samples. Similarly, for the Landscape+Into the Wild samples, 64.95\% of 16 participants rated our results higher than those by GlueGen (22.83\%) and AudioToken (12.23\%). Please check the supplementary material for the details of the user study.

\definecolor{pnpcolor}{HTML}{FEB5B1}
\begin{table}[!t]
\caption{\textbf{Quantitative Results for Image Editing}. We evaluate semantic consistency with AIS, AIC, and ISS metrics, and image quality with FID.
The highest scores are in bold, with second-best scores underlined. The text-based model is highlighted with a $\ddag$ symbol.}
\resizebox{0.98\linewidth}{!}{
\centering
\begin{tabular}{llcccc} 
\toprule
& Model & AIS $\uparrow$& AIC $\uparrow$& IIS $\uparrow$& 
 FID $\downarrow$ \\
 \midrule
 \multirow{4}{*}{\rotatebox[origin=c]{90}{L + ItW}} 
 & PnP~\cite{plugandplay}$^\ddag$ & -- & \textbf{.6833} & -- & 239.2\\
 & GlueGen~\cite{gluegen} & .5869 & .2254 & .6491  & 
 250.3 \\
 & AudioToken~\cite{audiotoken} & \underline{.6059} & .1849 & \underline{.7405} & 
 \underline{206.8}  \\
 & SonicDiffusion (Ours)  & \textbf{.6496}  & \underline{.3745} & \textbf{.7933}  & 
 \textbf{172.4}  \\
 \midrule
\multirow{4}{*}{\rotatebox[origin=c]{90}{GH}} 
 & PnP~\cite{plugandplay}$^\ddag$ & -- & \textbf{.6604} & -- & 
 \underline{199.5} \\
 & Gluegen~\cite{gluegen} & .4680 & .2275 & .5425 & 
 208.9  \\
 & AudioToken~\cite{audiotoken} & \underline{.4749} & .1450  & \underline{.5441}  & 
 206.8 \\
 & SonicDiffusion (Ours) & \textbf{.5491}  & \underline{.3575} & \textbf{.6255} &  
 \textbf{152.7} \\
 \bottomrule
 \end{tabular}
 }
 \label{tab:editing}
 \end{table}

We also show the ability of our model to control image editing outcomes through simple manipulations of input sounds. Our model can blend characteristics from two distinct audio sources by linearly interpolating their audio embeddings, as shown in Fig.~\ref{fig:sound_manipulation}(a). This results in outputs representing seamlessly transitions between features corresponding to each audio input. Additionally, as illustrated in Fig.~\ref{fig:sound_manipulation}(b), our model responds to changes in audio volume, allowing for precise manipulation of image content. The degree of change directly correlates with the volume adjustments, offering users detailed control over the audio's influence on the image.
\vspace{-0.05cm}
\section{Ablations}
To explore the influence of loss functions and training strategies on our SonicDiffusion model, we conducted an extensive series of ablation studies. These experiments aim to illuminate how different components and methodologies affect the model's efficacy in audio-visual alignment and overall performance.\\

\noindent\textbf{Impact of Loss Functions in the Training of SonicDiffusion's Audio Projector.}
In the first training stage of our model, we employ two loss functions: (i) contrastive loss and (ii) mean squared error (MSE) loss. We conducted ablation studies to understand the effects of omitting one of these losses. Table~\ref{tab:ablations_losses} presents the results of these experiments. When the contrastive loss is removed, our network struggles to distinguish between different audio conditionings. On the other hand, omitting the MSE loss results in the model still being able to differentiate between various audio cues, but with a longer convergence time and reduced accuracy. Including the MSE loss is particularly crucial as it facilitates the alignment of the audio space with the pre-trained CLIP space, though not through a direct mapping due to the simultaneous application of the contrastive loss. In essence, the contrastive loss adds another layer of complexity to this alignment. After completing the first stage of training, we continue to train the audio projector module but with a reduced learning rate. This step is vital as it enables the network to directly capture the semantic correlations between audio and image, enhancing the overall effectiveness of our SonicDiffusion model.\\

\begin{table}[!t]
\caption{\textbf{Impact of loss functions used in training Audio Projector on the performance of SonicDiffusion.} We report the performance using both contrastive and MSE losses (first row), using only the MSE loss (second row), and using only the contrastive loss (third row).}
\centering
\begin{tabular}{llllll}
\toprule
&   & AIS $\uparrow$& AIC $\uparrow$& IIS $\uparrow$& FID $\downarrow$ \\
\midrule
\multirow{3}{*}{\rotatebox[origin=c]{90}{L + ItW}} 
& SonicDiffusion& .7390 & .5436 & .8898 & 118.6 \\ 
& $\;$ w/o Contrastive Loss & .7245 & .3919 & .8432 & 121.6 \\ 
& $\;$ w/o MSE Loss& .6827 & .4018 & .7767 & 124.9\\
\midrule
\multirow{3}{*}{\rotatebox[origin=c]{90}{GH}} 
& SonicDiffusion& .6237 & .6050 & .7411 &  123.5  \\ 
& $\;$ w/o Contrastive Loss & .6560 & .3694 & .7264 & 133.4 \\ 
& $\;$ w/o MSE Loss& .6269 & .4027 & .6826 & 142.5 \\ 
\bottomrule
\end{tabular}
\label{tab:ablations_losses}
\end{table}

\noindent\textbf{Impact of Different Training Strategies on the Performance.}
Our goal is to establish a conditioning space that effectively guides the diffusion process. Leveraging embeddings from a pre-trained, well-represented space is a recognized strategy in diffusion models. Large text encoders such as CLIP~\cite{clip} and T5~\cite{T5} are commonly employed for this purpose due to their robust pre-trained capabilities. In our approach, we extract audio features from CLAP~\cite{clap} and consider an additional learning phase to map these features from the CLAP embedding domain to the CLIP embedding domain. To assess the impact of various training strategies on our model's performance, we conduct a series of experiments. These experiments focus on evaluating changes in performance under different conditions: representing audio features with a single token, mapping CLAP embeddings to CLIP space solely using diffusion loss (thereby omitting the first stage of training), and freezing the Audio Projector's parameters during the second stage of training. The outcomes of these experiments are summarized in Table~\ref{tab:ablations_train}. We found that relying exclusively on the diffusion loss for learning the mapping led to a notable decrease in the model's performance. Notably, the model seems to lose its ability to distinguish between different audio inputs. Freezing the Audio Projector does not consistently improve results and can even lower image quality, as indicated by the FID scores. Additionally, we also observed that representing audio with just a single token, as in AudioToken~\cite{audiotoken}, leads to worse performance.

\begin{table}[!t]
\caption{Impact of various training choices on the performance of our proposed SonicDiffusion model.}
\centering
\resizebox{0.998\linewidth}{!}{
\begin{tabular}{ll@{$\quad$}c@{$\quad$}c@{$\quad$}c@{$\quad$}c}
\toprule
& & AIS $\uparrow$& AIC $\uparrow$& IIS $\uparrow$& FID $\downarrow$ \\
\midrule
\multirow{3}{*}{\rotatebox[origin=c]{90}{L + ItW}} 
& SonicDiffusion& .7390 & .5436 & .8898 & 118.6  \\
& $\;$ w/o 1st Stage Training& .6013 & .0600 & .5160 & 116.7 \\
& $\;$ w/ Audio Proj. Frozen Stage 2& .7488 & .6020 & .8920 & 128.3  \\
& $\;$ w/ Single Token Audio Repr.& .7042 & .4618 & .8281 & 136.6 \\ 
\midrule \multirow{3}{*}{\rotatebox[origin=c]{90}{$\quad\;$GH}} 
& SonicDiffusion& .6237 & .6050 & .7411 & 123.5 \\
& $\;$ w/o 1st Stage Training& .4376 & .0888 & .5128 & 177.6  \\
& $\;$ w/ Audio Proj. Frozen Stage 2& .5849 & .5972 & .7188 & 128.3  \\
& $\;$ w/ Single Token Audio Repr.& .6863 & .4625 & .7647 & 119.0 \\ 
\bottomrule
\end{tabular}}
\label{tab:ablations_train}
\end{table}
\section{Limitations and Failure Cases}
\label{app:limitations}

\begin{figure}[!t]
    \centering
    \includegraphics[width=0.9\linewidth]{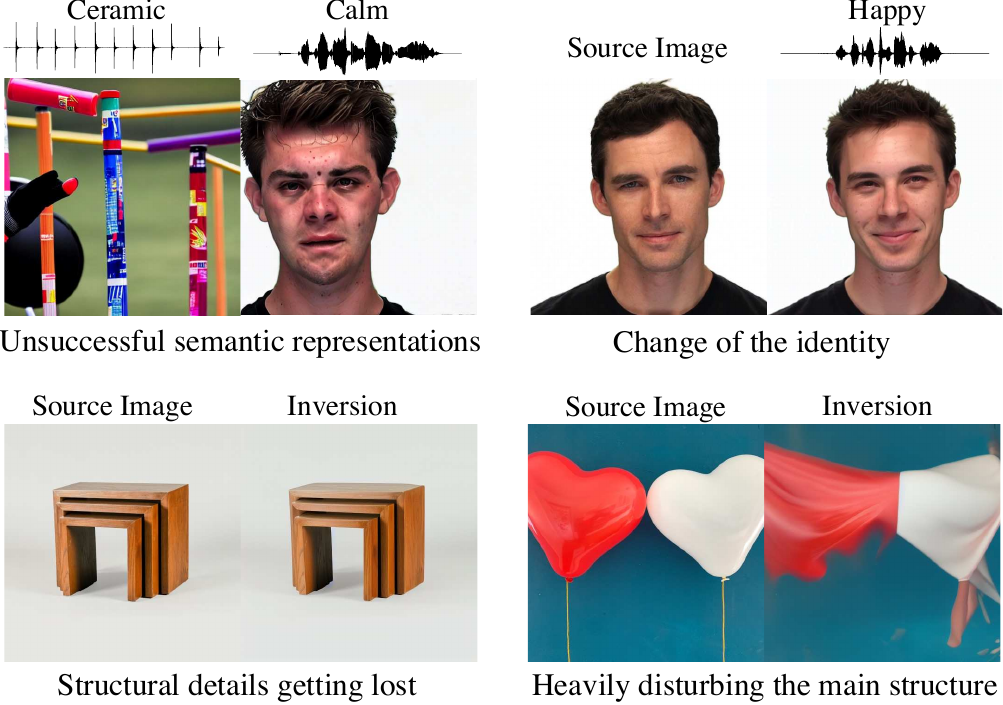}
    \caption{\textbf{Limitations of the SonicDiffusion model.} The figure illustrates various scenarios where the SonicDiffusion model does not adequately perform in the generation or modification of images based on audio cues. Issues arise from imprecise semantic interpretations of audio signals, as well as from artifacts introduced during the Stable Diffusion process. In the context of image editing, our model may inadvertently replace original content, conduct ineffective semantic modifications, or alter the subject's identity. Notably, some of the editing challenges are attributed to the use of DDIM inversion, which can result in the insertion of extraneous elements, the omission of fine details, or significant structural disruptions. }
    \label{fig:failure_cases}
\end{figure}

Our SonicDiffusion model demonstrates an encouraging capacity for image generation that aligns with the semantic context of provided audio input.  Nonetheless, as illustrated in Fig.~\ref{fig:failure_cases}, it is not without its shortcomings.  In particular, the model occasionally does not understand the essence of the audio content. One significant source of these suboptimal outcomes is the artifacts intrinsically produced by the underlying Stable Diffusion framework. This is evident when processing the Greatest Hits dataset, where the model's rendition of hands and objects in motion often falls short. Similarly, with the RAVDESS dataset, which primarily features human facial expressions, the model sometimes struggles with accurately rendering facial features, such as teeth, resulting in clearly noticeable anomalies. When it comes to image editing, the shortcomings become more pronounced with the employment of the DDIM inversion method. This technique is liable to obliterate delicate details in intricate scenes or integrating extraneous elements, thereby skewing the image's authentic structure. Instances of such editing impediments are exemplified in the third row of Fig.~\ref{fig:failure_cases}.

\section{Conclusion}
\label{sec:conclusion}
Our SonicDiffusion model introduces an original framework for sound-guided image generation and editing, leveraging the robustness of the SD model. We achieve seamless integration of audio modality with minimal training overhead by adapting cross-attention layers in SD to respond to audio prompts. Our method allows for the incorporation of audio context that aligns with the input sound, effectively enhancing both image generation and editing tasks. The results demonstrate our framework's strong performance in multimodal image synthesis, demonstrating significant advancements in the field of multimodal image synthesis.

\begin{acks}
This work has been partially supported by AI Fellowships to B. C. Biner provided by the KUIS AI Center, by BAGEP 2021 Award of the Science Academy to A. Erdem, by GEBIP 2018 Award of the Turkish Academy of Sciences, and by an Adobe research gift. 
\end{acks}

\bibliographystyle{ACM-Reference-Format}
\bibliography{bibliography.bib}

\end{document}


\title{Supplemental Material -- SonicDiffusion: Audio-Driven Image Generation and Editing with Pretrained Diffusion Models}

\author{Burak Can Biner}
\email{bbiner21@ku.edu.tr}
\orcid{}
\affiliation{%
  \institution{Koç University}
  \country{Turkey}
}

\author{Farrin Marouf Sofian}
\email{}
\orcid{}
\affiliation{%
  \institution{Koç University}
  \country{Turkey}
}

\author{Umur Berkay Karakaş}
\email{}
\orcid{}
\affiliation{%
  \institution{Koç University}
  \country{Turkey}
}

\author{Duygu Ceylan}
\email{duygu.ceylan@gmail.com}
\affiliation{%
  \institution{Adobe Research}
  \country{United Kingdom}
}

\author{Erkut Erdem}
\email{erkut@cs.hacettepe.edu.tr}
\orcid{0000-0002-6744-8614}
\affiliation{%
  \institution{Hacettepe University}
  \country{Turkey}
}

\author{Aykut Erdem}
\email{aerdem@ku.edu.tr}
\orcid{0000-0002-6280-8422}
\affiliation{%
  \institution{Koç University}
  \country{Turkey}
}

\renewcommand{\shortauthors}{Biner, et al.}

\maketitle

\renewcommand{\paragraph}[1]{{\vspace{1mm}\noindent \bf #1}}

 In this document, we aim to provide a deeper understanding and a more detailed description of our SonicDiffusion model, offering insights that complement the main paper. This includes a comprehensive breakdown of our training pipeline (Sec.~\ref{app:training-pipeline}), detailing the preprocessing step (Sec.~\ref{app:preprocessing}) and the two-stage process of initial alignment (Sec.~\ref{app:stage1}) and parameter-efficient tuning (Sec.~\ref{app:stage2}) to ensure clarity and reproducibility\footnote{We will also publicly share our code to facilitate further research.}. We then provide detailed descriptions of the evaluation metrics (Sec.\ref{app:metrics}) and the specifics of our user study (Sec.~\ref{app:user-study}), along with the comprehensive review of the competing approaches (Sec.~\ref{app:competing}). Moreover, we analyze the role of inserting gated cross-attention at different layers (Sec.~\ref{app:layers}), explain the choices made during inference (Sec.~\ref{app:inference}), and present additional results and visual comparisons to further illustrate the effectiveness of our approach (Sec.~\ref{app:additional-results}). Finally, we explain the broader impact of our work (Sec.~\ref{app:impact}). 
\vspace{4mm}

\section{Implementation and Training Details}

\subsection{Training Pipeline}
\label{app:training-pipeline}
The overall training pipeline of SonicDiffusion is illustrated in Fig.~\ref{fig:train_overview}. Our  model employs a two-stage training process, each stage serving a specific purpose in achieving audio-driven image synthesis. The initial stage is dedicated to training the audio projector module. The second stage of training focuses on integrating gated cross-attention layers within the Stable Diffusion (SD) model framework. Below, we discuss the details of data processing as well as the two-stage training framework.

\subsection{Preprocessing}
\label{app:preprocessing}
In SonicDiffusion, we employ the CLAP model~\cite{clap} as audio feature extractor. CLAP effectively merges audio and language into a single shared embedding space, showing strong performance with robust generalization capabilities across various downstream tasks, including sound event detection, acoustic scene classification and speech-based emotion recognition. We preprocess the audio input in accordance with CLAP. Specifically, we generate log-mel spectrograms with a hop size of 320 seconds, a window size of 1024 seconds, and 64 mel bins. Consistent with the training of Stable Diffusion v1.4, our model processes images of size 512$\times$512 pixels. We apply random horizontal flipping as our data augmentation technique. We conducted additional experiments with color space and geometric augmentations, such as color jitter, grayscale, random perspective, and random rotation, but observed that these led to degraded performance. Apart from horizontal flipping, the only data augmentation found to be beneficial is randomly cropping images instead of applying center cropping. To facilitate text-audio mapping as part of our initial training loss, we generated the caption of each image in training set using BLIP~v2~\cite{blipv2}. We also tested audio captioning models in~\cite{ac, act} but found that BLIP v2 provided captions which are much better aligned with the audio and image pairs. %

\subsection{Training of The Audio Projector}
\label{app:stage1}
The audio projector module in our framework is designed to transform audio clips into semantically rich tokens. Specifically, it projects each input audio clip into $K$ tokens with $C$ channels. For our model, we chose $K=77$ and $C=768$, precisely aligning with the text conditioned embedding space of Stable Diffusion. This alignment allows us to implement a straightforward Mean Squared Error (MSE) loss between corresponding text and audio tokens. In addition to MSE loss, we calculate a contrastive loss. As discussed in the main paper, we select two audio clips from the same class, $\mathbf{a}_0$ and $\mathbf{a}_1$, and $N$ audio clips from different classes, and calculate the contrastive loss as follows:
\begin{equation}
   \mathcal{L}_\text{InfoNCE} = -\log \frac{\exp(\langle \mathbf{a}_{0},\mathbf{a}_{1}\rangle)}{\sum_{i=1}^{N+1} \exp(\langle \mathbf{a}_{0},\mathbf{a}_{i}\rangle) }\;,
\end{equation}
where $N$ is set to 128 in our experiments.

We calculate both the MSE and the contrastive losses for each of the $K$ audio tokens. Drawing inspiration from GlueGen~\cite{gluegen}, which showed that information in CLIP text tokens is not uniformly distributed, instead of averaging the loss across all the tokens, we apply a weighted summation. This is formulated using reverse sigmoid weighting, as given below: 
\begin{equation}
    \mathcal{L} = \sum_{i=1}^{N} w_i \mathcal{L}_\text{i}\;, \quad w_i = \frac{t}{t + \exp(i/t)}\;,
\end{equation}
where $\mathcal{L}_\text{i}$ is the total loss calculated for token $i$, and $t$ is temperature coefficient, which is set to 5 in our experiments.

It is also worth noting is that keeping $C$ at 768 enables us to initialize the weights of our audio-image cross-attention layers from the pre-trained text-image cross-attention layers of Stable Diffusion. This initialization strategy is helpful for faster convergence. 

\subsection{Audio-conditioned Tuning of Stable Diffusion}
\label{app:stage2}
In the training of the Stable Diffusion model, encoders like CLIP are typically kept frozen. However, in our model's second stage of training, we keep the audio projector module trainable to further enhance its robustness. Initially, the audio projector focuses on class separation and aligning audio features with those of the image captions. By continuing the training in the second stage, we aim to improve its ability to match image and audio features. It should be noted that we perform the second stage training using a lower learning rate of 1e-5 to preserve the initial knowledge learned in the first stage.

Various approaches have been explored in the literature for aligning different domains or modalities with the conditioning space of Stable Diffusion (SD). Uni-ControlNet~\cite{unicontrol-net} and GlueGen~\cite{gluegen} append new modality information to text tokens, while AudioToken~\cite{audiotoken} learns a specific token representing the unique information provided in the input. These methods use the existing text-image cross attention layers of SD, originally trained for text conditioning, to provide conditioning with respect to the new modalities. Our approach introduces a new trainable cross-attention layer, which allows a new modality to develop its own learned representation during image generation.

In our training scheme, the total number of trainable parameters is 34M, with 8M for the audio projector, and 26M for the gated cross-attention layers. This is considerably low as compared to the original UNet architecture, which has a total of 880M parameters. We further initialize the weights of the new audio-image cross attention layers with those of the text-image cross attention layers. Hence, our model achieves efficient tuning to introduce the ability to condition the network with respect to a new modality. This efficiency extends beyond model parameter storage. Another notable advantage of using adapter layers is their ability to prevent catastrophic forgetting in the pre-trained model. In our experiments, we validate this by demonstrating the use of mixed modality inputs including audio along with text prompts.

To enable classifier-free guidance, we set the CLAP embedding to zero with ten percent probability during the tuning stage. Similar to null-text embedding,
we also create a ``null audio embedding". 
Followingly, we define the classifier-free guidance as follows: 
\begin{equation}
    \epsilon = w\epsilon_{\theta} (\mathbf{x}, t, \mathbf{a}) - (1-w)\epsilon_{\theta} (\mathbf{x}, t_{\emptyset}, \mathbf{a}_{\emptyset})
\end{equation}
where $t$ is the text, $t_{\emptyset}$ is the null text, $a$ is the audio, $a_{\emptyset}$ is the null audio embedding and $w$ is the CFG scale.

\section{Evaluation Setup}

\subsection{Evaluation Metrics}
\label{app:metrics}
In our analysis, we consider four separate evaluation metrics: Audio-Image Similarity (AIS)~\cite{audiotoken}, Image-Image Similarity (IIS)~\cite{audiotoken}, and Audio-Image Content (AIC)~\cite{audiotoken} to assess the semantic relevance, and FID metric~\cite{frechet_inception_distance} to evaluate the sample quality. In the following, we provide the detailed descriptions of these metrics.

\paragraph{Audio-Image Similarity (AIS)} is designed to measure the semantic similarity between the audio input and the generated image, determining how well the image reflects the audio content. In particular, this similarity is estimated by employing the Wav2CLIP model~\cite{wav2clip}, a model providing a common semantic space for images and audio like CLAP~\cite{clap} does for text and audio. As noted in~\cite{audiotoken}, relying solely on the similarity score between features of the input audio and the features of the generated image can be misleading due to variations in the scales of the similarity scores. Hence, AIS not only compares the generated image with its corresponding input audio but also contrasts it with audio samples from the entire validation set. Specifically, the similarity of the generated image both with the conditioning audio clip as well as other audio clips in the validation set are computed. Then the validation audio clips which result in a lower similarity then the conditioning audio clip are identified. The ratio of such audio clips to the total number of audio clips is reported as the AIS score. Thus, AIS serves as a reference-based score and provides a more comprehensive and meaningful measure of similarity.

\paragraph{Image-Image Similarity (IIS)} is designed to quantify the semantic similarity between a generated image and the corresponding ground truth image linked to the input audio. It serves to evaluate how accurately the generated image captures the semantic content of the target image. The scoring method employed is similar to that of AIS. The similarity score is calculated using the CLIP~\cite{clip} model, comparing the generated image with both its ground truth image and image samples from the validation dataset. The IIS score for a model is then computed as an average of these similarity scores across all entries in the validation set, providing a holistic measure of the model's performance.

\paragraph{Audio-Image Content (AIC)} is designed to evaluate the relevance of the content of a generated image to its corresponding ground-truth audio label. It assesses the alignment between the class predicted by an image classifier and the actual audio label. Specifically, we use CLIP as a zero-shot classifier to compute the probability of the image belonging to each of the class labels in the dataset. An agreement is noted when the ground-truth label achieves the highest probability among these. The AIC score is then calculated as the average of these agreement instances.

\paragraph{Fr\'{e}chet Inception Distance (FID)} is used to assess the perceptual quality and diversity of the generated images. In particular, it measures the photorealism by measuring the difference between the distribution of the real images and that of the generated or manipulated images using deep features.

\begin{figure*}[!t]
\begin{tabular}{c@{}c}
\includegraphics[width=0.5\linewidth]{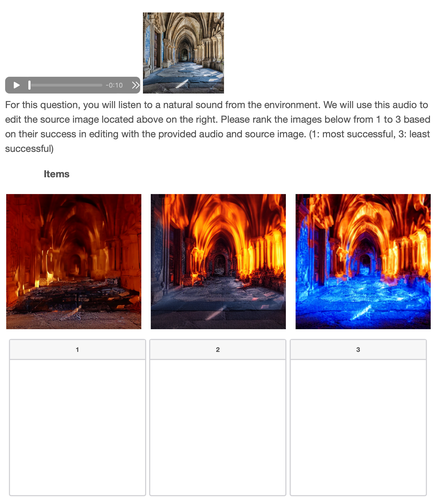} &
\includegraphics[width=0.5\linewidth]{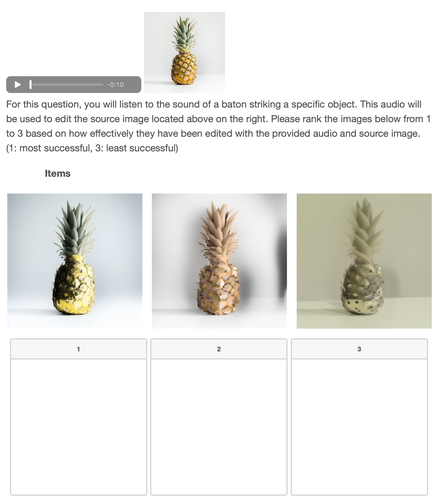} \\
{\small Landscape + Into the Wild} &
{\small Greatest Hits}
\end{tabular}
  \caption{\textbf{User study GUI}. Screenshots of two sample questions from the user study where the participants are asked to rank the model outputs in terms of editing quality with respect to provided audio and the source image.}
  \label{fig:user-study}
\end{figure*}

\subsection{User Study}
\label{app:user-study}
To subjectively evaluate the effectiveness of SonicDiffusion, we performed a user study using the Qualtrics platform. We picked a total of 43 source images (20 for the model trained on Greatest Hits, and 23 for the model trained on Landscape + Into the Wild) from Unsplash website~\cite{unsplash} and audio inputs to generate the audio-conditioned edited versions of these images using our approach and the competing Glue-Gen~\cite{gluegen} and AudioToken~\cite{audiotoken} models. In each question, the participants are asked to rank the outputs of each model according to editing quality. Screenshots of two sample questions from the user study are given in Fig.~\ref{fig:user-study}.

\subsection{Competing Approaches}
\label{app:competing}
In our evaluation, we compare our SonicDiffusion model with seven state-of-the-art methods, SGSIM~\cite{sgsim}, Sound2Scene~\cite{sound2visual}, GlueGen~\cite{gluegen}, ImageBind~\cite{imagebind}, CoDi~\cite{codi}, AudioToken~\cite{audiotoken}, TempoTokens~\cite{yariv2023diverse}. Among these, SGSIM and Sound2Scene utilize GANs for audio-to-image synthesis, while GlueGen, CoDi and AudioToken integrate Stable Diffusion in their image generation pipeline. ImageBind employs a DALLE$\cdot$2-based generator~\cite{dalle2}. TempoTokens differs from the aforementioned models as it is designed for generating videos from audio clips, again leveraging Stable Diffusion. Hence, we consider the center frames of the generated videos when evaluating TempoTokens. We finetune GlueGen, AudioToken and Sound2Scene on our datasets. However, for CoDi and ImageBind, we use the pre-trained models provided by the authors due to their large-scale nature. We do not perform finetuning for TempoTokens, as it was already trained on the Landscape dataset. For SGSIM, we do not finetune their audio encoder but train StyleGAN2-ADA~\cite{Karras2020ada} on our datasets to run inference. To apply AudioToken and GlueGen for image editing, we extend their generative frameworks with PnP injection \cite{plugandplay} at self-attention layers 4-11 and residual layer 4. By injecting the features, we can successfully preserve the source image structure during the generation process. Below we provide training details of the aforementioned competing approaches.

\paragraph{SGSIM}. We first trained a separate StyleGAN2-ADA model for each of the three datasets using adaptive augmentation. For Landscape + Into-the-Wild, the model is trained for around 800K steps. For Greatest Hits and RAVDESS, their corresponding models have converged around 25K steps and 40K steps so we use these checkpoints. %

\paragraph{AudioToken}. We finetuned their embedder and Lora weights with all of our datasets. Although finetuning was not available within their provided codebase, we managed to add this feature. During finetuning, we used the parameters suggested in their GitHub repository for training. For Greatest Hits and RAVDESS, we finetuned for 60K steps rather than 30K as we observed that the models finetuned with 30K steps resulted in worse performance than the models finetuned with 60K steps. We used same parameters during generation and editing and applied injection at the same layers during editing.

\paragraph{Sound2Scene}. We used default parameters in their training script and finetuned their model for 100 epochs on each dataset. As their pre-trained generator was trained for 128$\times$128 images, we generated 128$\times$128 images for comparison with our model.

\paragraph{GlueGen}. We finetuned the GlueNet model on audios from all three datasets for 30 epochs, with a learning rate of 2e-5 and kept the rest of the parameters as the suggested defaults.

\paragraph{ImageBind}. We did not perform finetuning on ImageBind. We demonstrated generation results of ImageBind's shared embedding space using Stable Diffusion UnCLIP\footnote{https://github.com/Zeqiang-Lai/Anything2Image}. 

\paragraph{CoDi.} We did not perform finetuning on CoDi. We used the official repository and checkpoints to have audio to image generation results.

\section{Further Analysis}

\subsection{Impact of Where to Insert Gated Cross-Attention Layers}
\label{app:layers}
Stable-Diffusion v1.4 UNet configuration has sixteen text-image cross attention layers. Six of them lies inside encoder side, one of them lies inside the middle block, and nine of them lies inside the decoder side within blocks of 3 to 11. In SonicDiffusion, we inject gated cross-attention layers only into the middle block and decoder blocks. We perform an analysis on this design choice and investigate the impact of where to insert gated cross-attention layers within our SonicDiffusion model. Table \ref{tab:ablations_layers} summarizes the results of this analysis. As can be seen, injecting gated cross attention layers to only for the sixth to eleventh blocks of the UNet significantly diminishes the model's effectiveness, whereas injecting them only to the encoder blocks gives only slight improvements. 

\begin{table}[!h]
\caption{Influence of where to inject gated cross-attention layers within our SonicDiffusion model on model performance.}
\centering
\begin{tabular}{llllll}
\toprule

& Model & AIS $\uparrow$& AIC $\uparrow$& IIS $\uparrow$& FID $\downarrow$\\
\midrule
\multirow{3}{*}{\rotatebox[origin=c]{90}{L + ItW}} &
SonicDiffusion & .7390 & .5436 & .8898 & 118.6\\
& $\;$ w/ injection to layers 6-11& .7089 & .4242 & .8479 & 145.4\\
& $\;$ w/ injection to all layers& .7471 & .5676 & .8931 & 125.6\\
\midrule
\multirow{3}{*}{\rotatebox[origin=c]{90}{GT}} &
SonicDiffusion& .6237 & .6050 & .7411 & 123.5 \\
& $\;$ w/ injection to layers 6-11& .5968 & .4027 & .6638 & 160.5 \\
& $\;$ w/ injection to all layers& .6148 & .6250 & .7542 & 132.5 \\
\bottomrule
\end{tabular}
\label{tab:ablations_layers}
\end{table}

\subsection{Inference Time Choices}
\label{app:inference}
\begin{figure}[!t]
\centering
  \includegraphics[width=\linewidth]{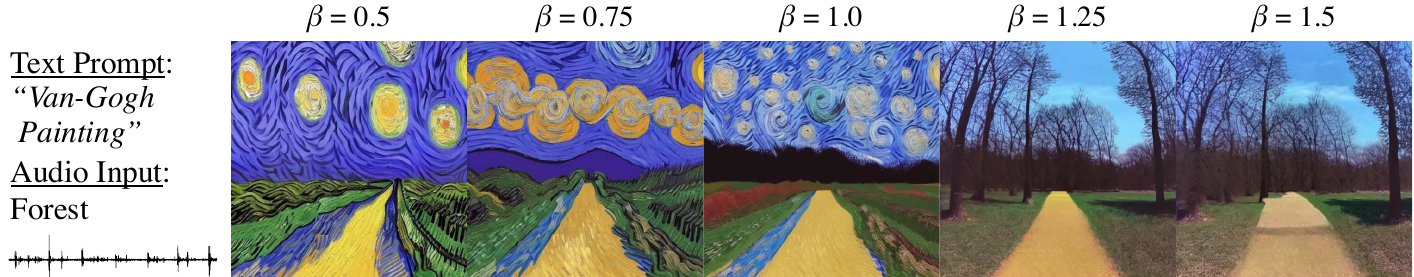}\vspace{-0.2cm}
  \caption{\textbf{Influence of the $\boldsymbol{\beta}$ coefficient.} As shown in this example, setting the value of $\beta$ specifies the strength of how the generated process will be modulated in relation to the audio versus text input.}
  \label{fig:text_audio_betas}
\end{figure}

Diffusion-based image generation offers extensive controllability over various aspects of image synthesis -- an attribute our SonicDiffusion model seeks to expand upon. In diffusion models, the classifier-free guidance (CFG) mechanism is used to modulate the influence of the conditioning factor. In our model, we harness CFG to finely balance the influence of both audio and text inputs. Additionally, we achieve finer control over these two modalities. 
As detailed in Equation 9, adjusting the $\beta$ value enables us to modulate the relative importance of audio versus text in the generation process (Fig.~\ref{fig:text_audio_betas}). Interestingly, the $\beta$ value can be varied across different layers of the UNet architecture, providing layer-specific control, which we haven't fully explored.

\begin{figure}[!b]
\centering
\begin{tabular}{c@{}c@{}c}
\includegraphics[width=0.31\linewidth]{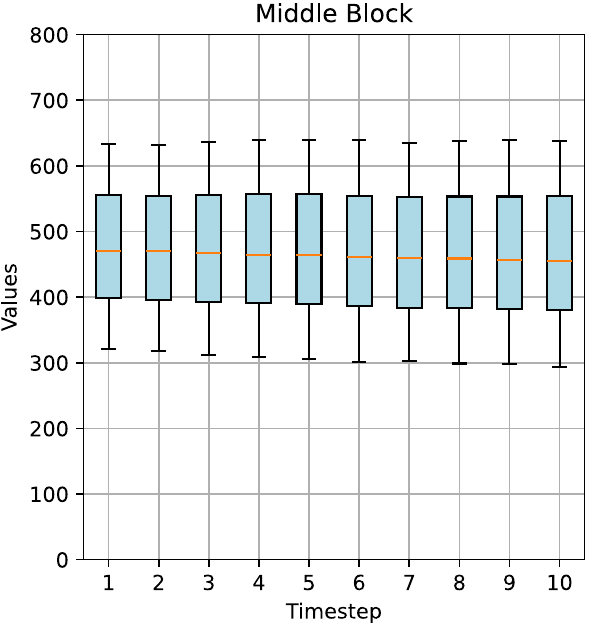} &
\includegraphics[width=0.31\linewidth]{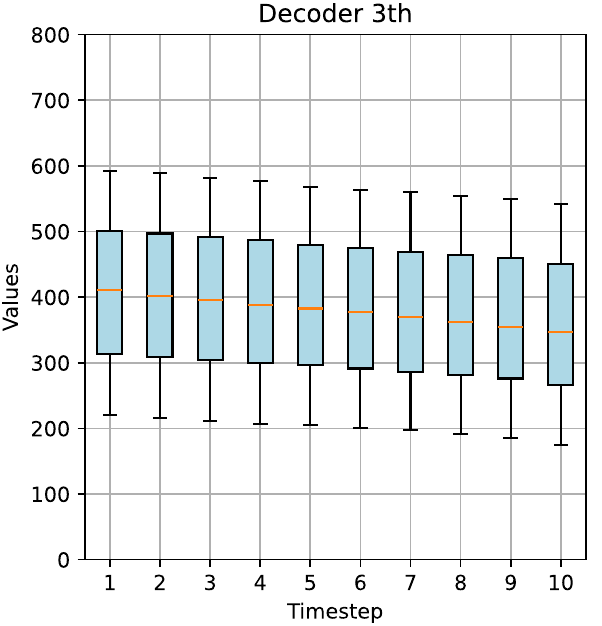} &
\includegraphics[width=0.31\linewidth]{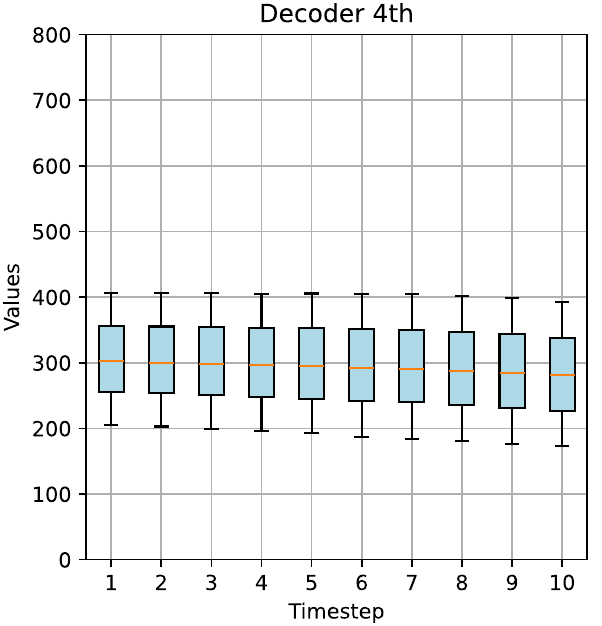}\\
\includegraphics[width=0.31\linewidth]{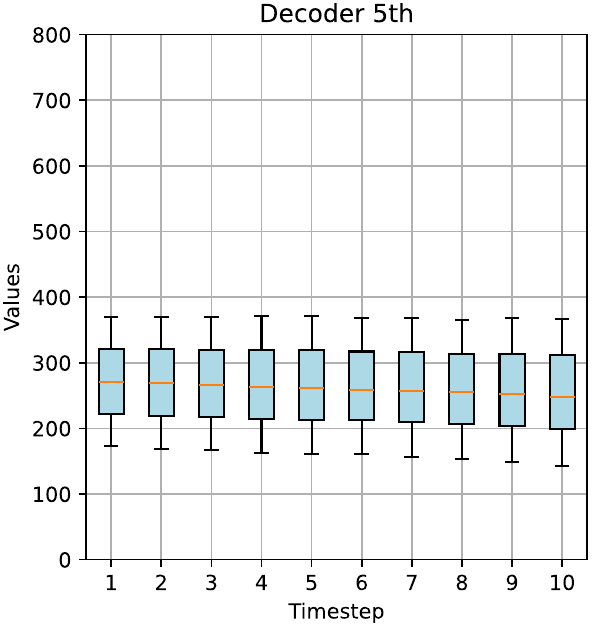}&
\includegraphics[width=0.31\linewidth]{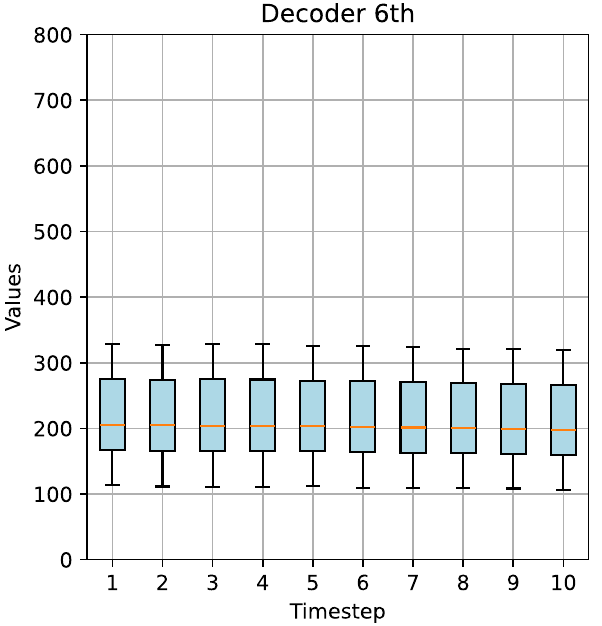} &
\includegraphics[width=0.31\linewidth]{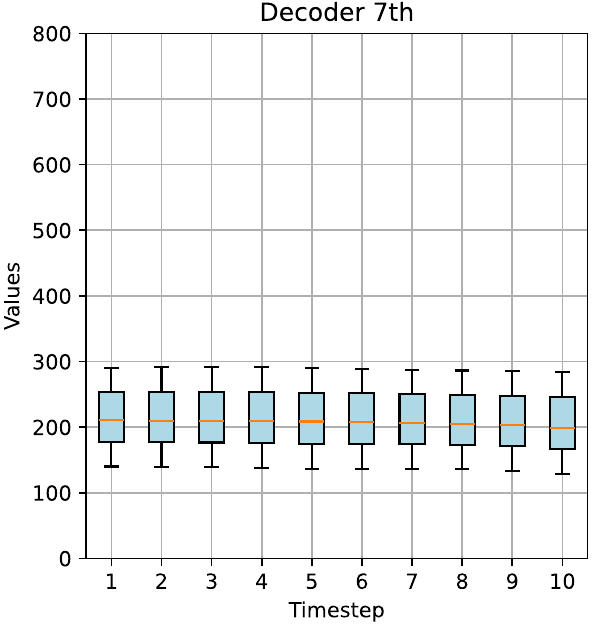}\\
\includegraphics[width=0.31\linewidth]{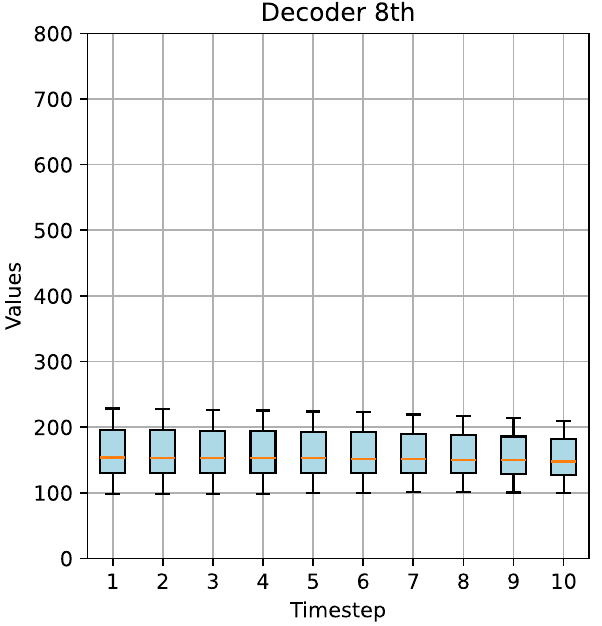} &
\includegraphics[width=0.31\linewidth]{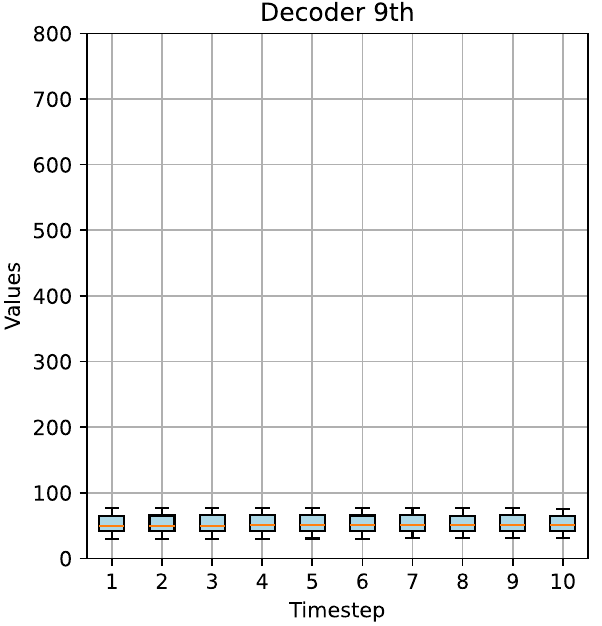} &
\includegraphics[width=0.31\linewidth]{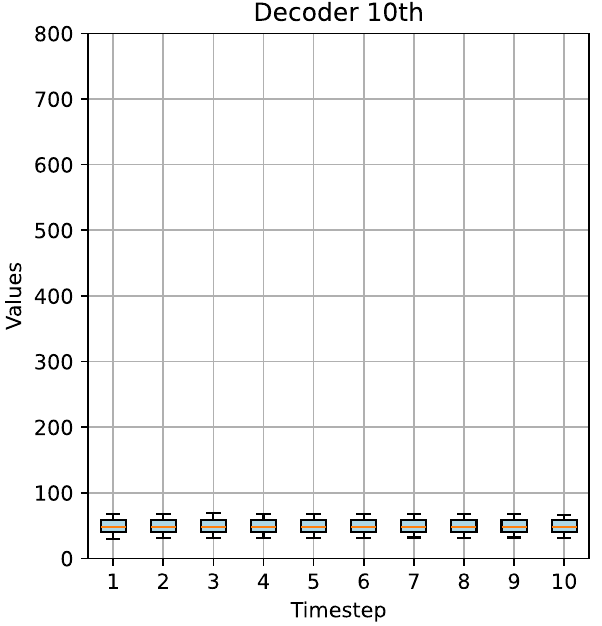}\\
\includegraphics[width=0.31\linewidth]{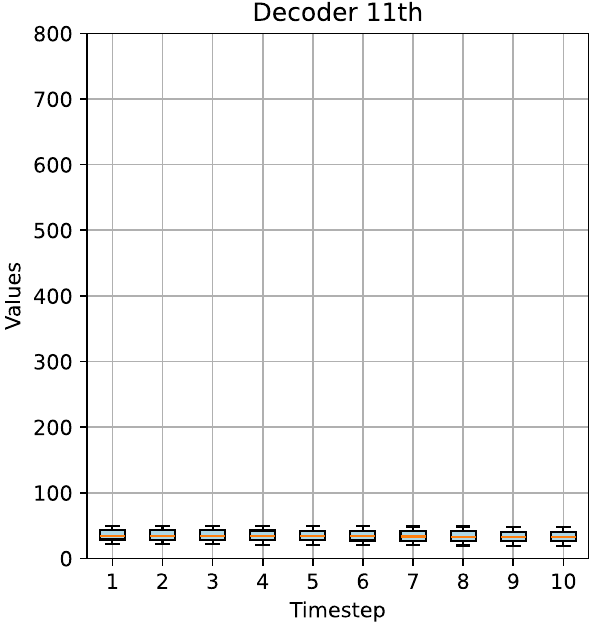}\vspace{-0.3cm}
\end{tabular}
\caption{\textbf{Norms of gated cross attention outputs across UNet layers}. Plots reveals the introduction of low-frequency information in early layers at the initial time steps where as high-frequency details are encoded in higher layers.}
\label{fig:adapter_norm}
\end{figure}

Previous studies, such as \citet{plugandplay}, have identified that the earlier layers of the UNet decoder are responsible for low-frequency image information, while the higher layers manage high-frequency details. Furthermore, \citet{patashnik2023localizing, balaji2022eDiff-I} show that initial steps of the diffusion process heavily influence the general layout of the generated image. We observe similar phenomenons for our injected gated cross-attention layers. Fig.~\ref{fig:adapter_norm} depicts box plots illustrating the norms of 150 generated images using our adapter blocks injected at different decoder layers along the diffusion steps. The analysis of norms within the middle block and Layers 3, 4, and 5 indicates that information of low frequency is primarily introduced during the initial time steps. This suggests that the input audio mostly guides the formation of coarse structures.
This layer-specific variation indicates that these layers are crucial for conveying different types of information during the generation process.

In the methodology we follow for image editing, the layers that we choose to inject features and the timesteps that we inject features are also design choices. We compare our results with other baselines using self attention layers 4-11 and residual layers 4 to have standard layers selected in the original PnP paper. However, some of the results are better when we use more residual injections like 4-11 self-attention and 4-6 residual blocks. This helps to preserve the original image structure better.

\subsection{Additional Results}
\label{app:additional-results}

\begin{figure*}[!t]
  \includegraphics[width=0.86\linewidth]{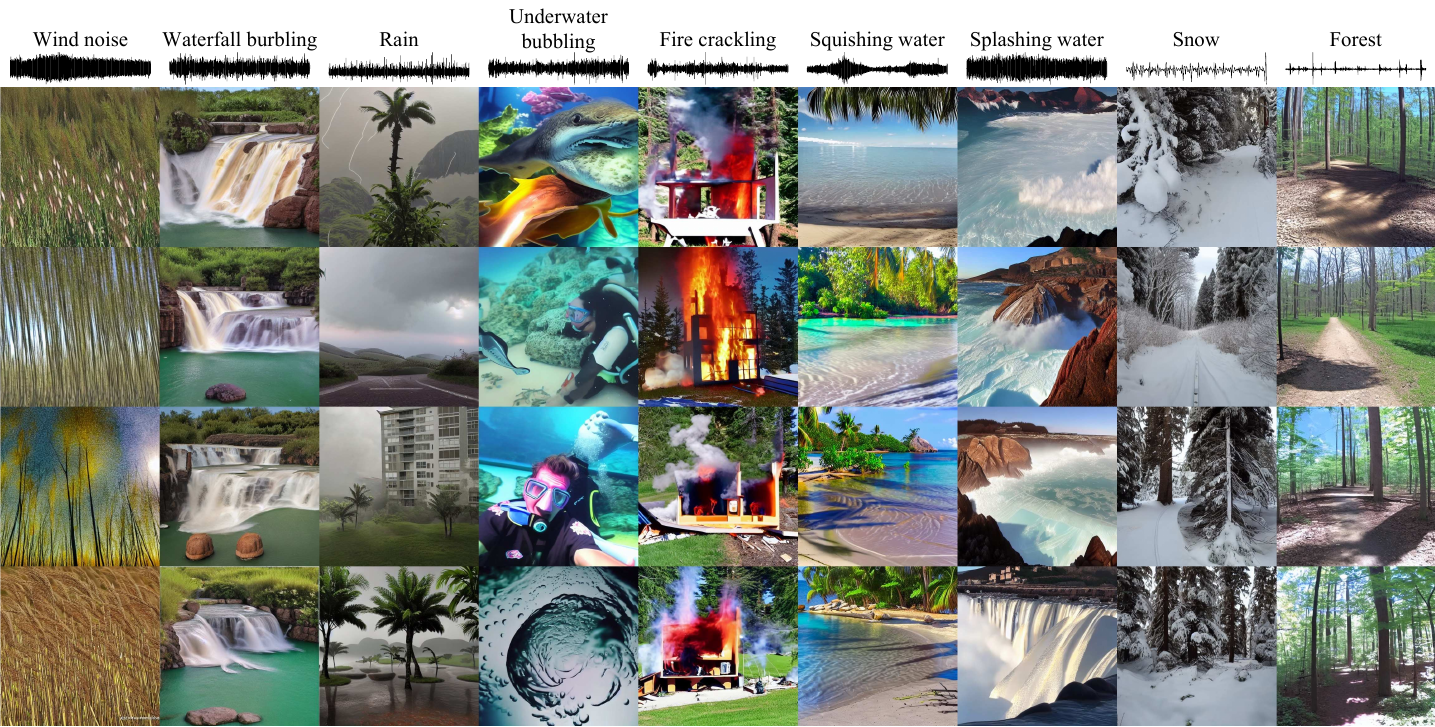}
  \caption{Additional audio-driven image generation results obtained with different seeds and audio clips from the Landscape + Into the Wild dataset.}
  \label{grid_landscape_gen}
\end{figure*}

\begin{figure*}[!t]
  \centering
  \includegraphics[width=0.86\linewidth]{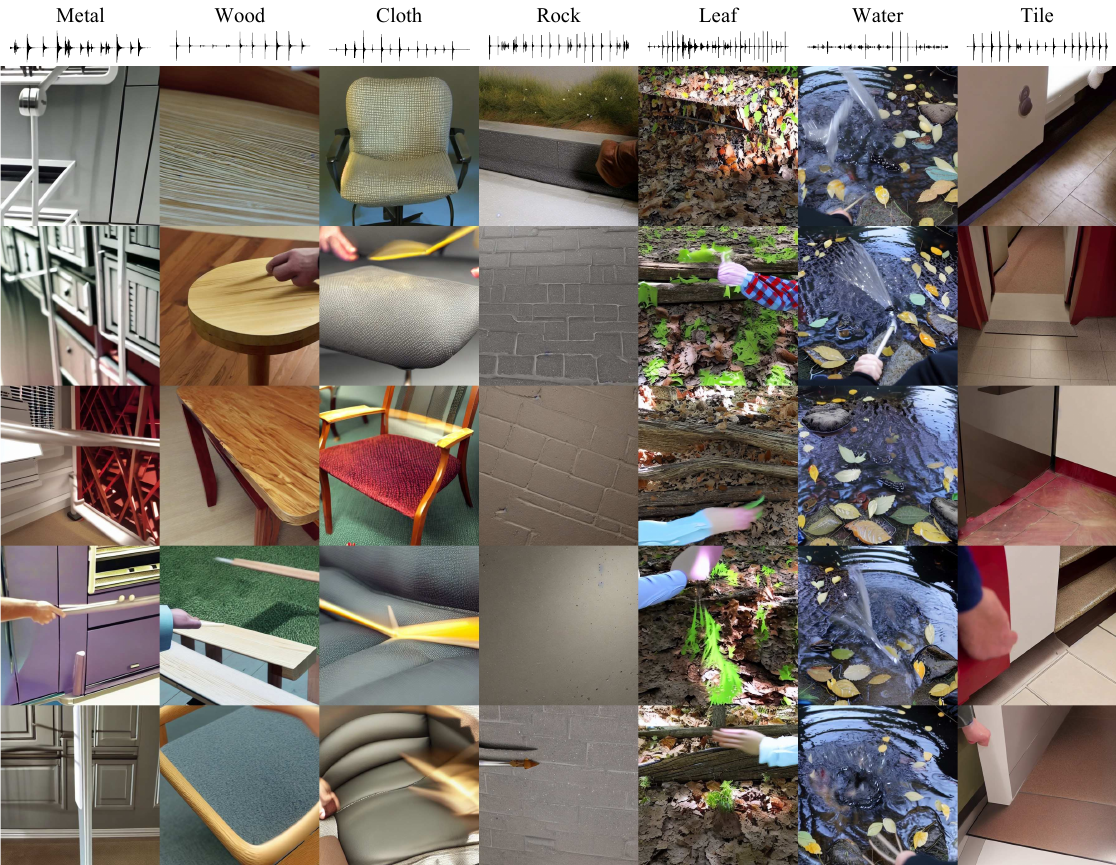}
  \caption{Additional audio-driven image generation results obtained with different seeds and audio clips from the Greatest Hits dataset.}
  \label{grid_gh_gen}
\end{figure*}

\begin{figure*}[!t]
  \includegraphics[width=0.8\linewidth]{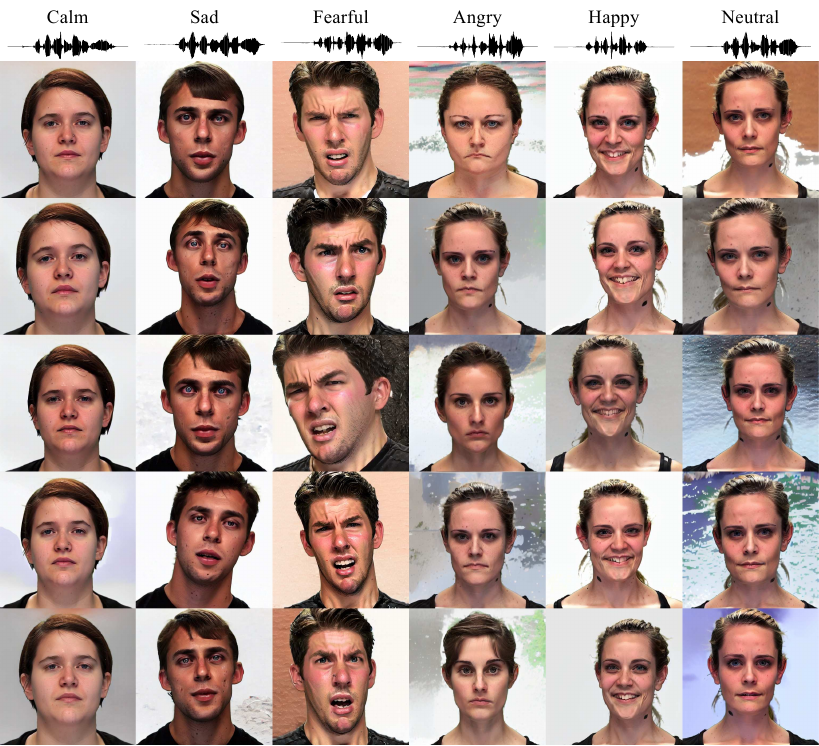}
  \caption{Additional audio-driven image generation results obtained with different seeds and audio clips from the RAVDESS dataset.}
  \label{grid_ravdess_gen}
\end{figure*}

\begin{figure*}[!t]
  \includegraphics[width=\linewidth]{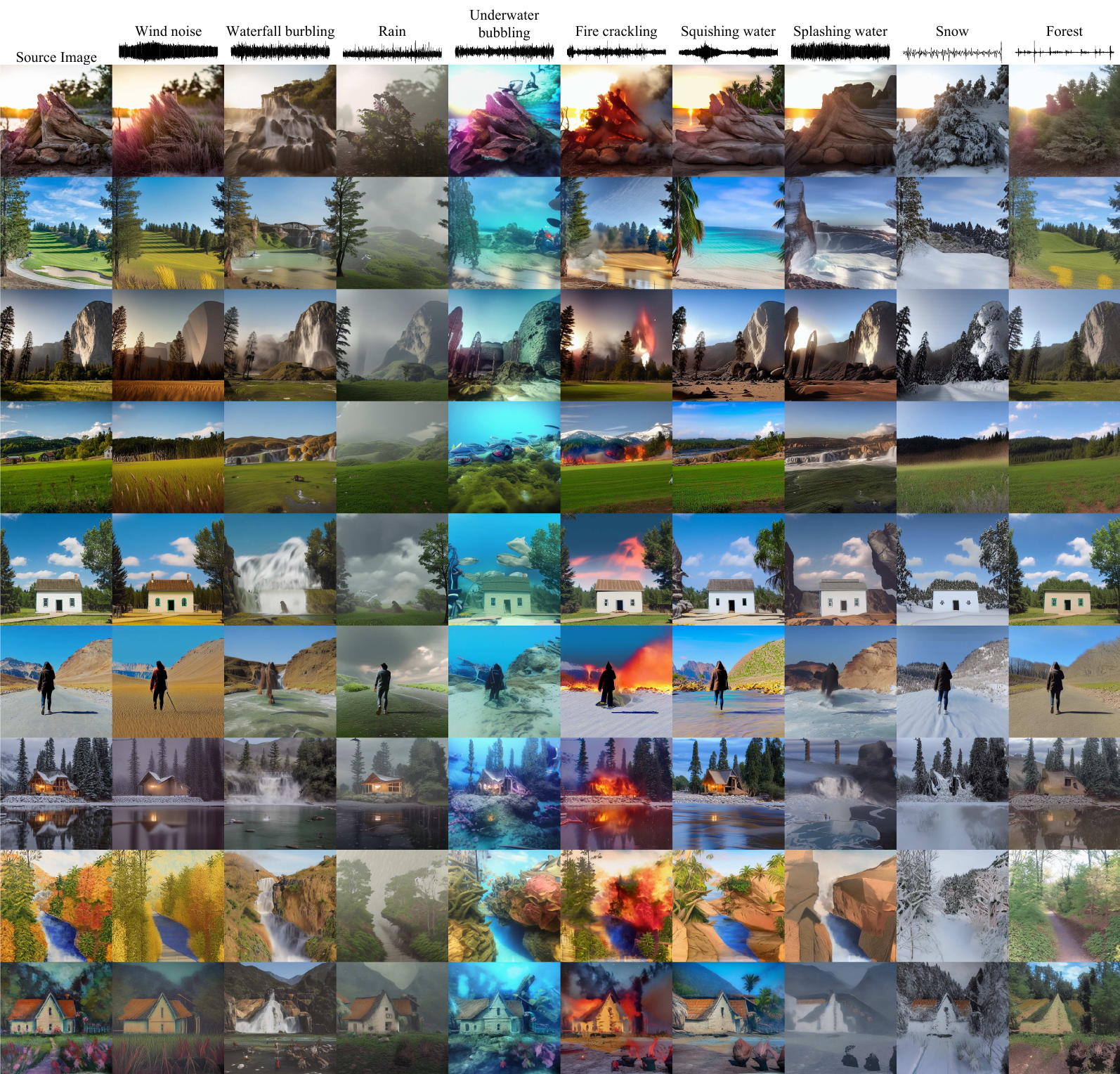}
  \caption{Additional audio-guided image editing resulting on a variety of source images influenced by different audio clips from the Landscape + Into the Wild dataset.}
  \label{grid_landscape_edit}
\end{figure*}

\begin{figure*}[!t]
  \centering
  \includegraphics[width=0.95\linewidth]{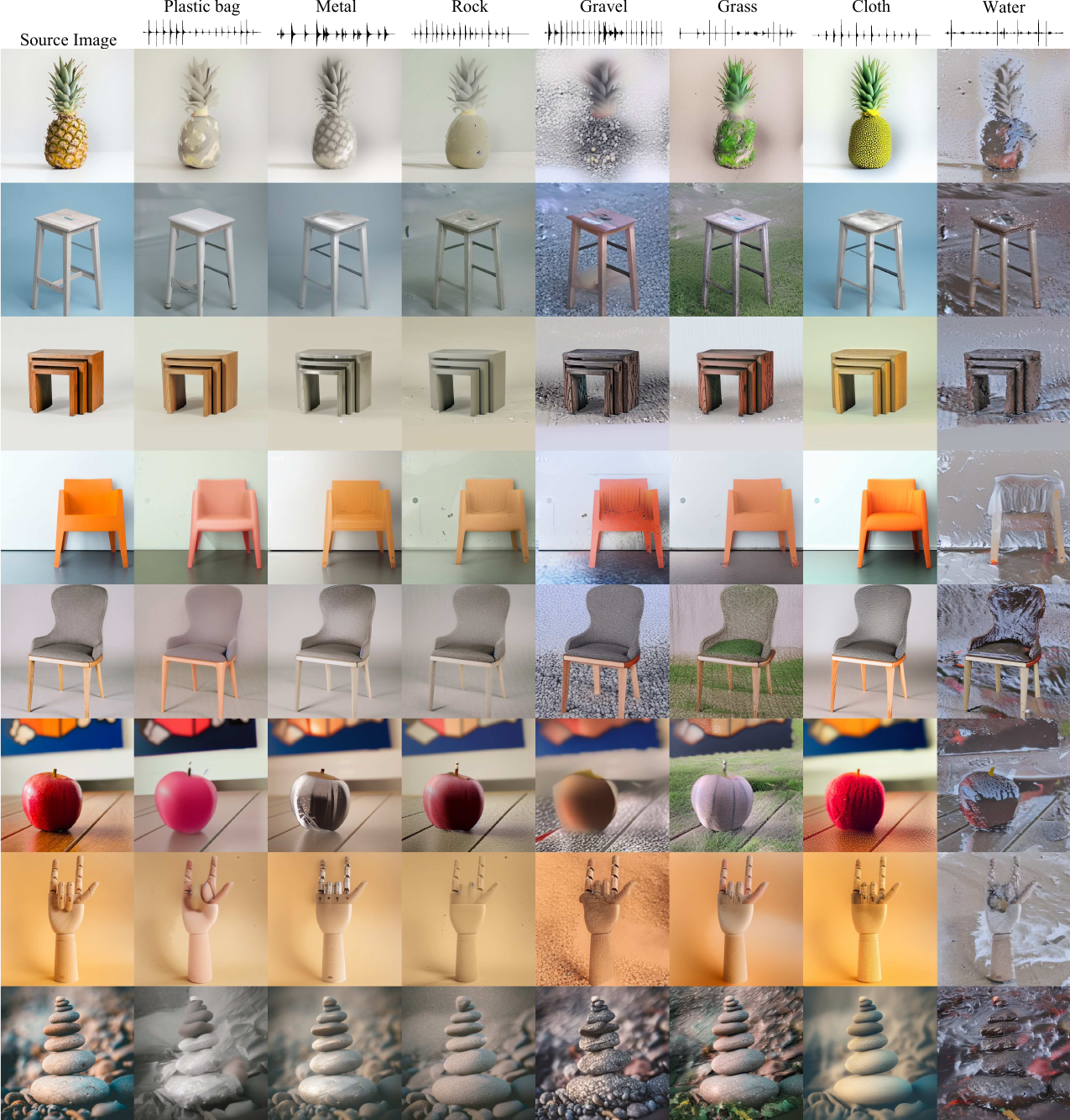}
  \caption{Additional audio-guided image editing results on a variety of source images influenced by different audio clips from the Greatest Hits dataset.}
  \label{grid_gh_edit}
\end{figure*}

\begin{figure*}[!t]
  \includegraphics[width=0.9\linewidth]{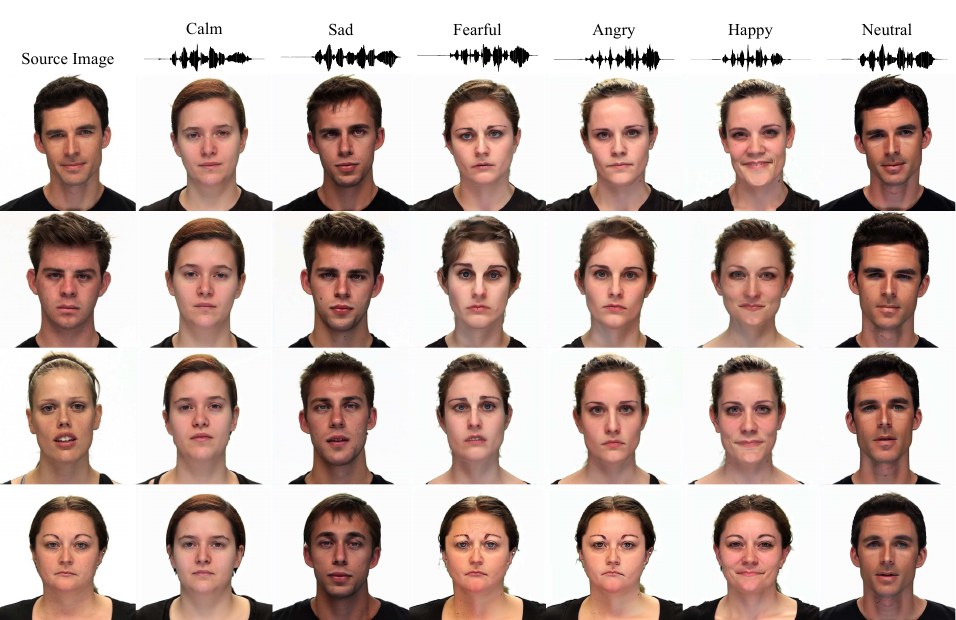}
  \caption{Additional audio-guided image editing results on a variety of source images influenced by different audio clips from the RAVDESS dataset.}
  \label{grid_ravdess_edit}
\end{figure*}

\begin{figure*}[!h]
    \centering
    \includegraphics[width=\linewidth]{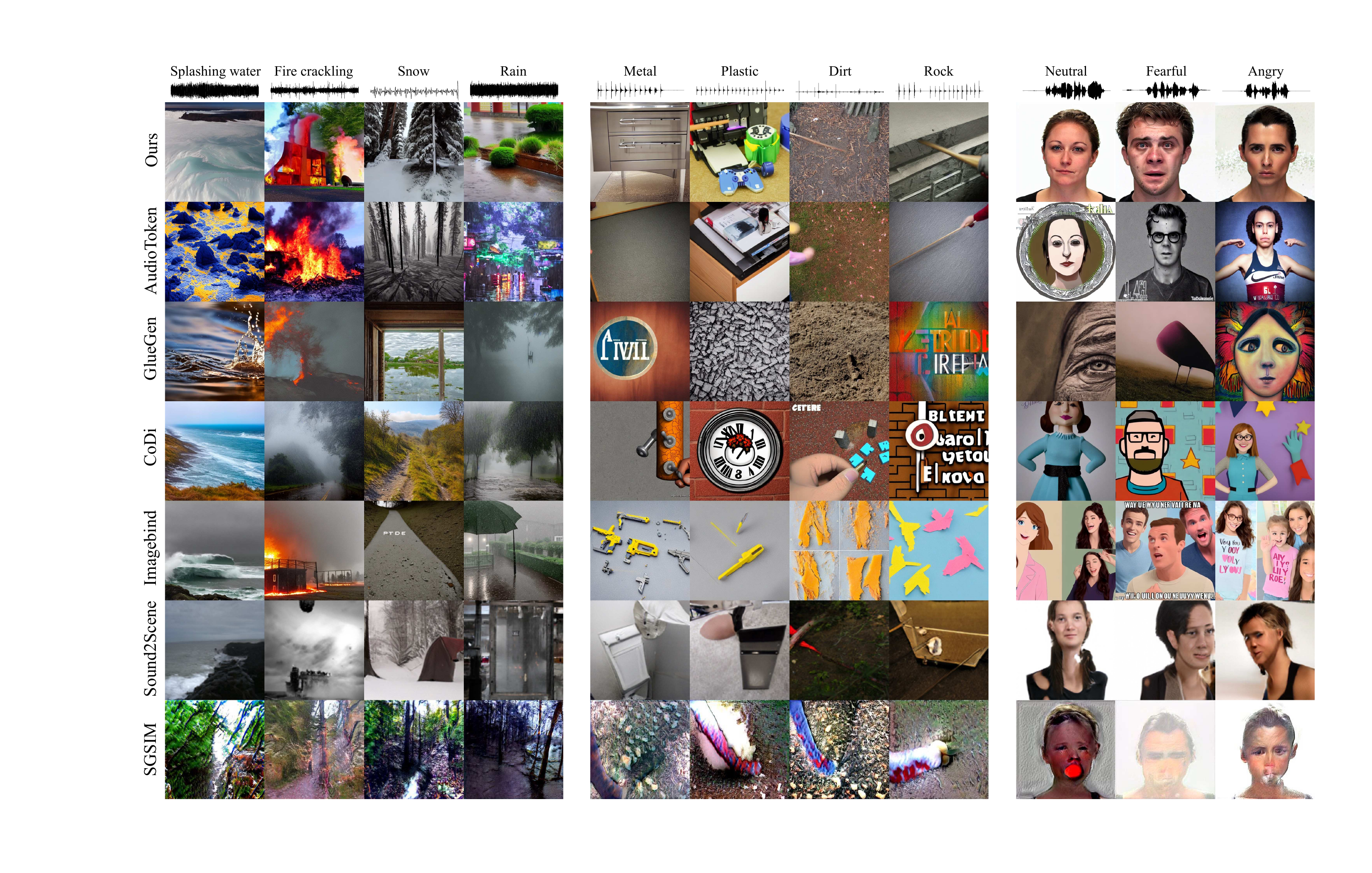}
    \caption{Additional comparisons against the current state-of-the-art audio-driven image generation approaches.}
    \label{fig:baseline_gen_compare}
\end{figure*}

\begin{figure*}[!t]
    \centering
    \includegraphics[width=\linewidth]{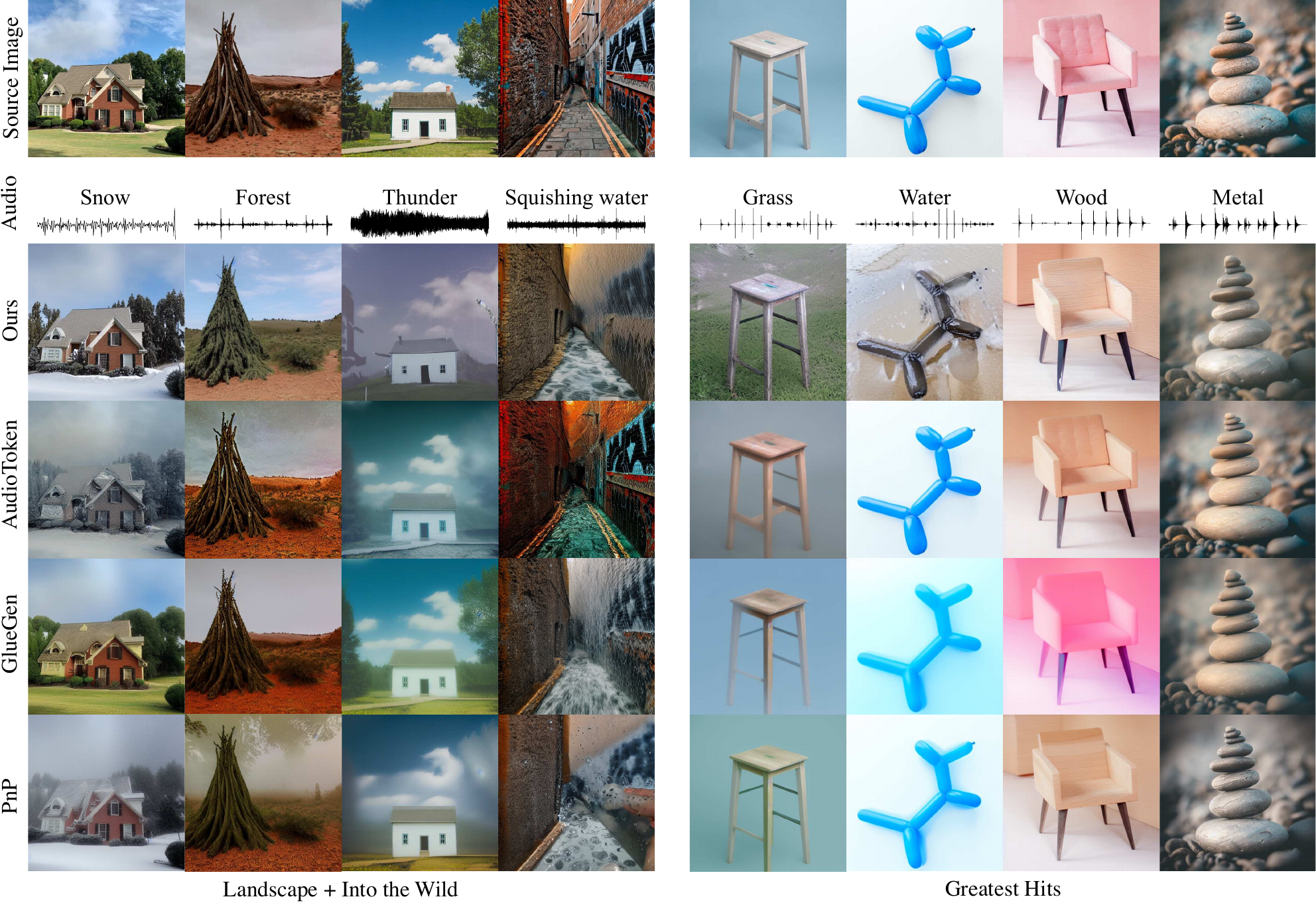}
    \caption{Additional comparsions against the current state-of-the-art audio-driven image manipulation approaches. }
    \label{fig:baseline_edit_compare}
\end{figure*}

\begin{figure*}[h!]
    \centering
    \begin{subfigure}{\textwidth}
\includegraphics[width=\textwidth]{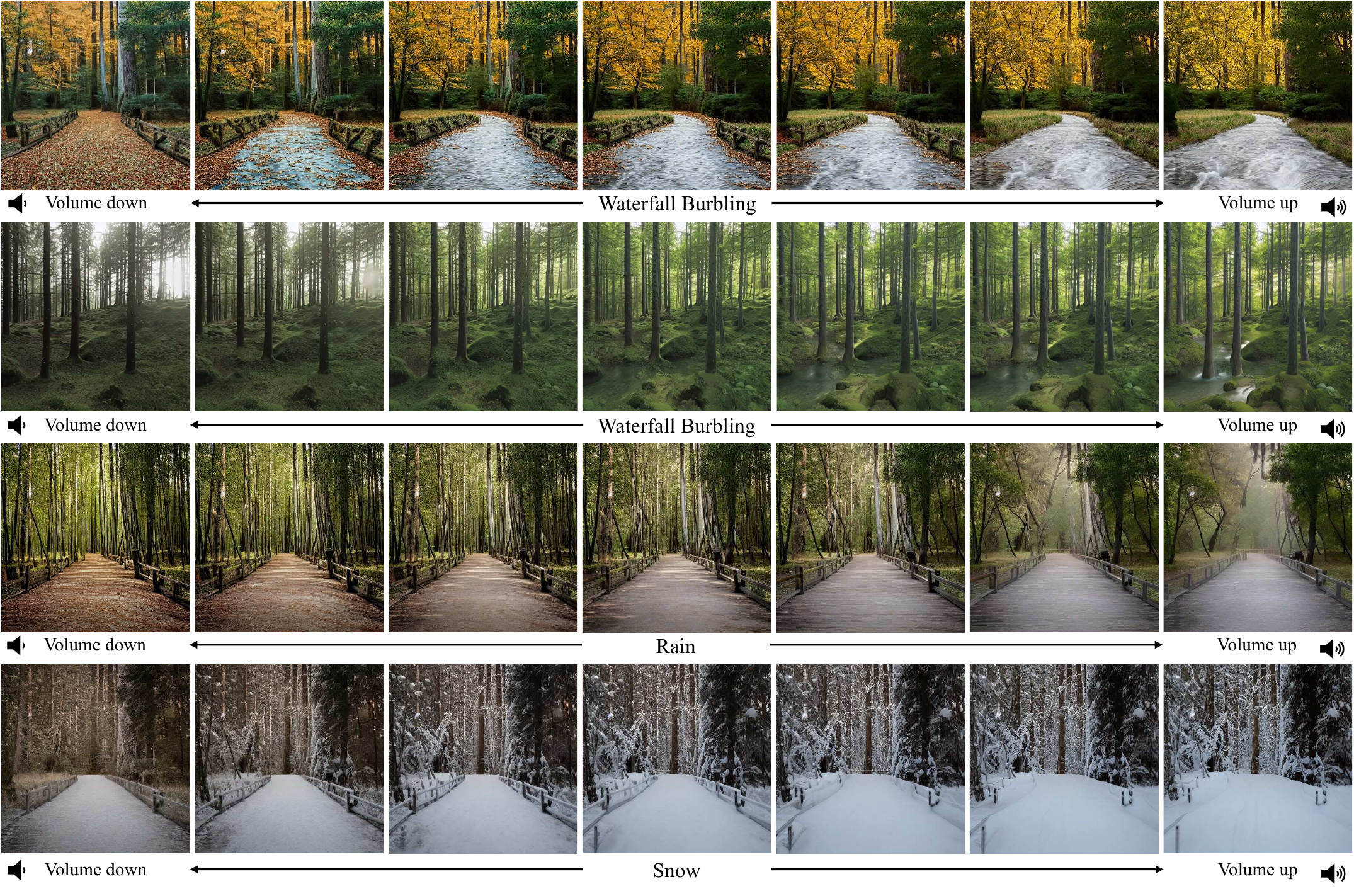}\vspace{-0.3cm}
        \caption{Volume Changes\vspace{0.3cm}}
        \label{fig:audio}
    \end{subfigure}

    \begin{subfigure}{\textwidth}    \includegraphics[width=\textwidth]{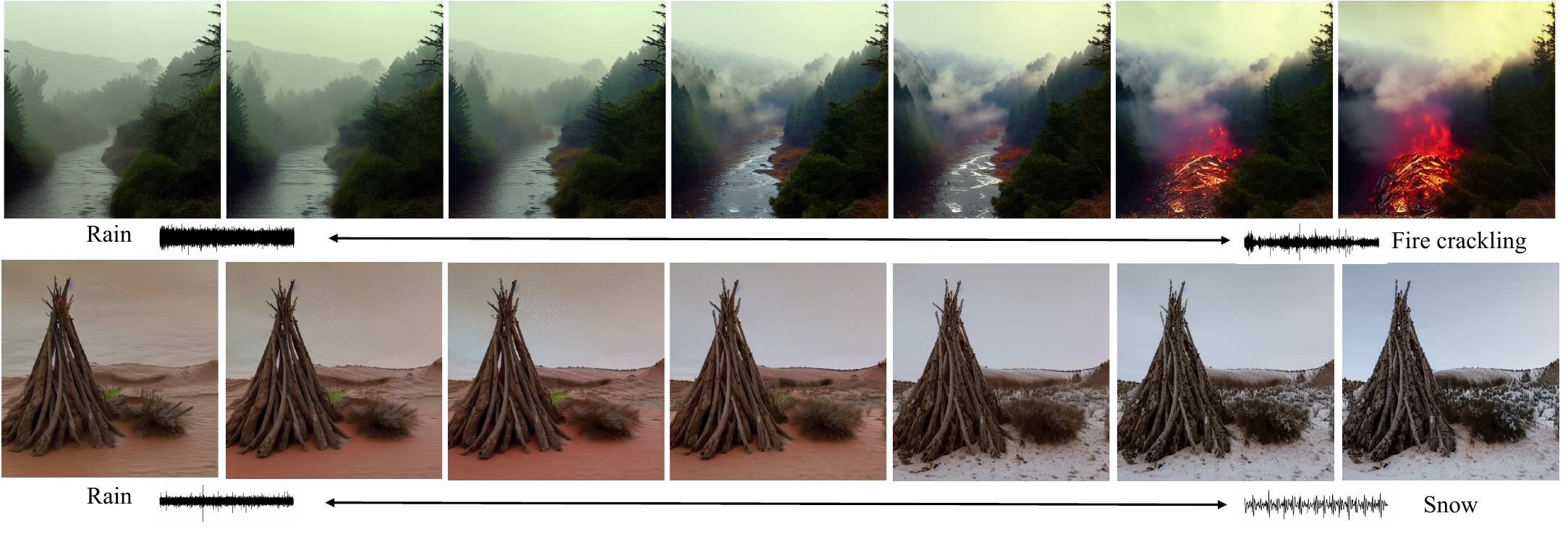}\vspace{-0.3cm}
        \caption{Sound Interpolation}    \label{fig:interpolations}
    \end{subfigure}
    \caption{Audio Interpolation and Mixing Impact on Image Editing: The effectiveness of the editing process is illustrated by (a) applying linear interpolation between two audio tracks, as shown in the figure, and (b) adjusting the intensity of the edits by manipulating the sound volume.}
    \label{fig:combined}
\end{figure*}

\begin{figure*}[!t]
    \centering
    \includegraphics[width=0.75\linewidth]{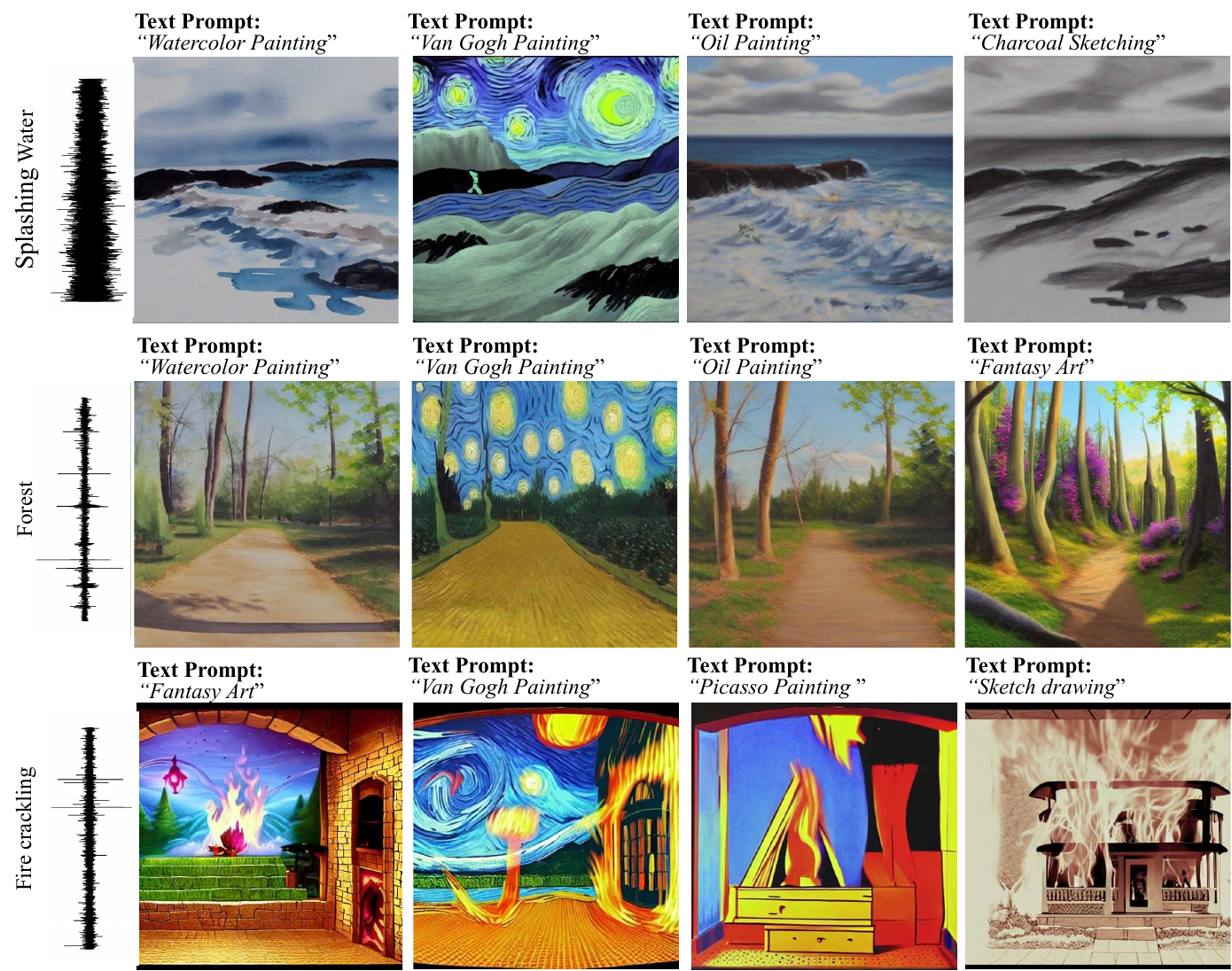}
    \caption{The integration of additional text prompts, representing different image styles, changes the overall appearance of the generated images, yet preserves the visual content conveyed by the audio inputs.}
    \label{fig:audio-f-text}
\end{figure*}

\begin{figure*}[!t]
    \centering
    \includegraphics[width=0.95\linewidth]{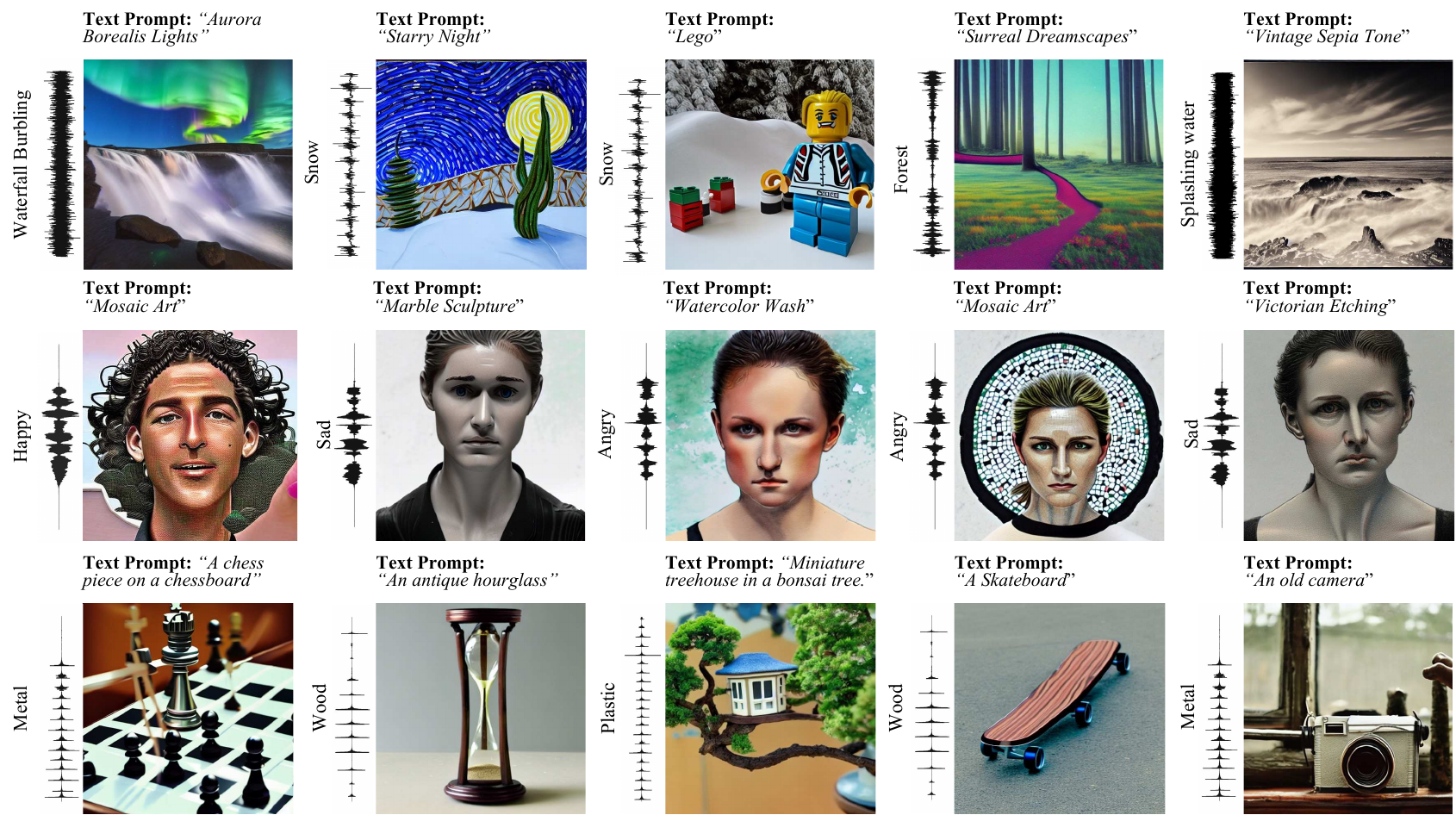}
    \caption{Sample outputs of the proposed SonicDiffusion model with mixed modality inputs. These images demonstrate our model's capability to synthesize novel images by harmonizing audio and text inputs, showing its adaptability across various text prompts and audio tracks.}
    \label{fig:audio-text-grid}
\end{figure*}

\begin{figure*}[!t]
    \centering
  \includegraphics[width=\linewidth]{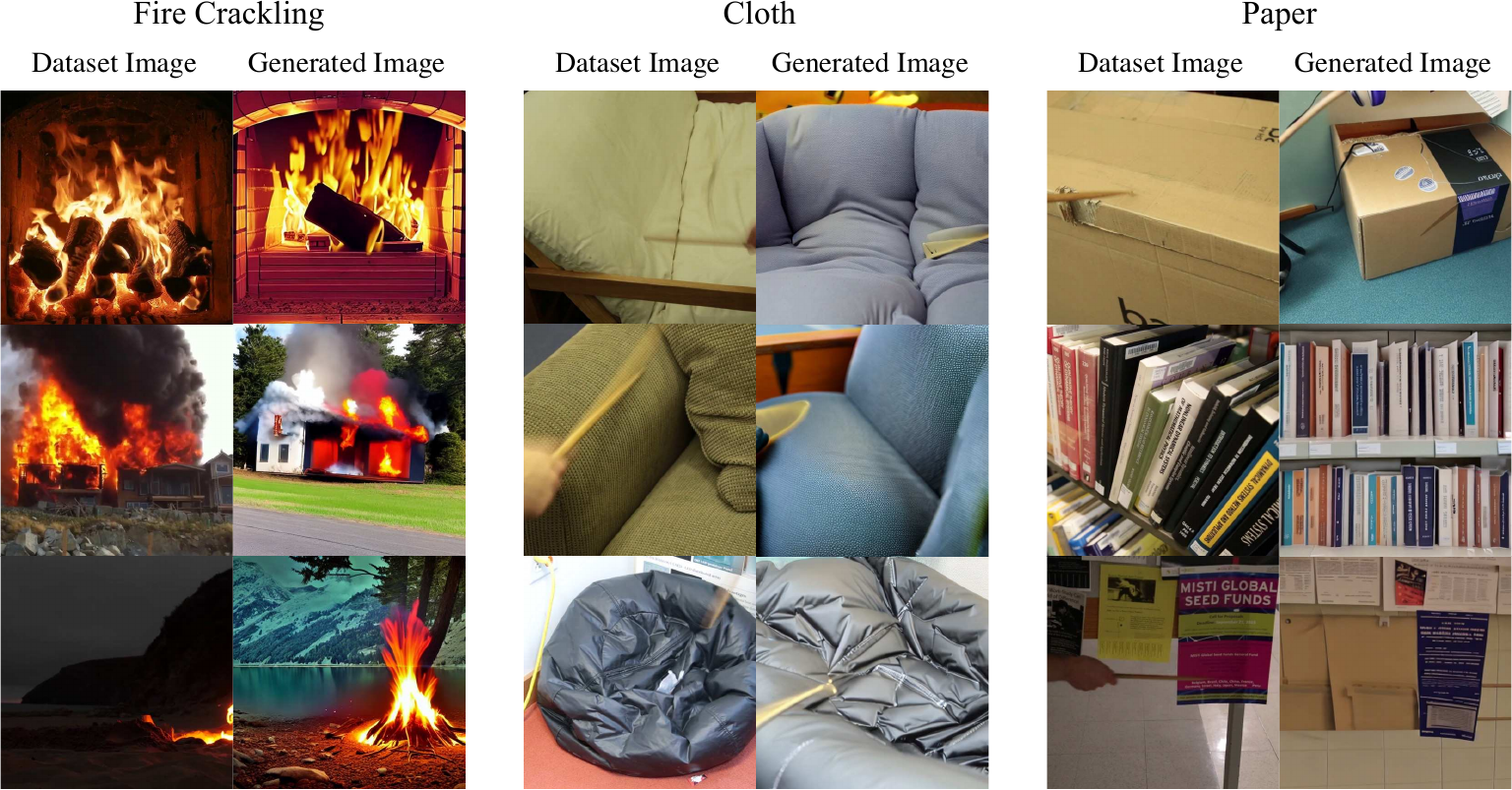}
    \caption{
    SonicDiffusion can extract detailed information from provided audio clips, effectively distinguishing between semantic meanings embedded within audio cues of the same category. Above, the images generated from sample input audio clips and those dataset images associated with these audio inputs are given side-by-side. Our model  identifies and depicts inherent auditory characteristics of various fire cracklings, cloths, and papers, revealing their unique acoustic signatures in the visual domain.
    }
    \label{fig:nuanced_audio}
\end{figure*}

In this section, we provide additional qualitative results obtained with our proposed SonicDiffusion model and further comparisons against the competing approaches. Fig.~\ref{grid_landscape_gen},~\ref{grid_gh_gen},~\ref{grid_ravdess_gen} present audio-driven image generation results using audio inputs from the Landscape + Into the Wild, Greatest Hits and RAVDESS datasets, respectively. Furthermore, we demonstrate the audio-driven image editing capabilities of our SonicDiffusion model in Fig.~\ref{grid_landscape_edit},~\ref{grid_gh_edit},~\ref{grid_ravdess_edit}, each utilizing audio inputs from the Landscape + Into the Wild, the Greatest Hits, and the RAVDESS datasets, correspondingly. In Fig.~\ref{fig:baseline_gen_compare} and Fig.~\ref{fig:baseline_edit_compare}, we provide qualitative comparisons of our SonicDiffusion against the state-of-the-art audio-driven image generation and audio-guided image editing models, respectively. Additionally, Fig.~\ref{fig:combined} demonstrated the dynamic control our model offers in editing results through simple manipulations of input sounds, such as adjusting audio volume and blending two different audio sources via linear interpolation of their embeddings. Fig.~\ref{fig:audio-f-text} and Fig.~\ref{fig:audio-text-grid} show sample images generated using mixed modality inputs that blend both text and audio, illustrating the versatility of our SonicDiffusion model in handling multimodal data. Lastly, Fig.~\ref{fig:nuanced_audio} shows our model's ability to generate novel images by juxtaposing the images produced from some input audio clips against those associated with these audio inputs, illustrating the semantic diversity inherent in the audio.

\section{Broader Impact}
\label{app:impact}
The broader impact of our SonicDiffusion model spans several domains, reflecting a significant step forward in the integration of audio signals for image synthesis and editing. This work opens new possibilities for creating visual content that resonates with the emotional and semantic nuances of sound, offering novel applications in digital art, media production, and virtual reality. For instance, our model can enhance the accessibility of visual media for those with hearing impairments by translating sound into corresponding visual narratives. Additionally, it paves the way for more intuitive human-computer interaction where users can guide visual creation through voice and sound, making technology more accessible to those without expertise in complex image editing software.

Moreover, our method holds potential for advancing the field of automated content generation, where audio tracks can influence the generation of dynamic visual scenarios, such as in video games or immersive simulations. It also contributes to the research community by providing a new direction for multimodal learning, encouraging further exploration into the interplay between different sensory inputs and computational creativity.

However, it is also essential to acknowledge and address potential ethical considerations, such as the use of this technology to create deepfakes or other forms of misleading content. Responsible usage guidelines and further research into detection and prevention mechanisms are crucial to ensure the positive impact of this technology on society.

\bibliographystyle{ACM-Reference-Format}
\bibliography{bibliography.bib}